\documentclass{article}

\usepackage[final,main]{neurips_2026}

\makeatletter
\renewcommand{\@notice}{}
\makeatother

\usepackage[utf8]{inputenc}
\usepackage[T1]{fontenc}
\usepackage{hyperref}
\usepackage{url}
\usepackage{booktabs}
\usepackage{amsfonts}
\usepackage{amsmath}
\usepackage{amssymb}
\usepackage{nicefrac}
\usepackage{microtype}
\usepackage{graphicx}
\usepackage{xcolor}
\usepackage{subcaption}
\usepackage{bm}
\usepackage{multirow}

\DeclareMathOperator*{\argmax}{arg\,max}
\DeclareMathOperator*{\argmin}{arg\,min}

\title{Measuring Black-Box Confidence via Reasoning Trajectories:\\Geometry, Coverage, and Verbalization}

\author{%
  \normalfont
  Marc Boubnovski Martell$^{1}$ \quad
  Josefa Lia Stoisser$^{1}$ \quad
  Kaspar M\"artens$^{1}$ \quad
  Jialin Yu$^{2}$ \\
  Robert Kitchen$^{1}$ \quad
  Philip Torr$^{2}$ \quad
  Jesper Ferkinghoff-Borg$^{1}$ \\[4pt]
  \small $^{1}$Novo Nordisk \quad $^{2}$University of Oxford
}

\begin{document}

\maketitle

\begin{abstract}
Reliable confidence estimation gates safe deployment of chain-of-thought (CoT) reasoning through text-only APIs, yet the dominant black-box baseline, self-consistency over $K$ samples, is linearly expensive and ignores the geometry of the trace. We introduce a black-box trajectory-confidence score that embeds a CoT as a sliding-window trajectory and measures its convergence toward external answer anchors with a one-parameter softmax, requiring no logits, hidden states, or supervised calibrators. On six $(\text{benchmark}, \text{reasoner})$ settings over MedQA-USMLE, GPQA Diamond, and MMLU-Pro $\times$ \{Gemini~3.1~Pro, Claude Sonnet~4.6\}, fusing this score with coverage and verbalized-confidence channels at $K{=}4$ Pareto-improves self-consistency at $K{=}8$ in $6/6$ settings (median AUC $0.78$ vs.\ $0.71$, $\Delta\mathrm{AUC}=+0.075$); a fixed-pick control ($+0.060$) and an E5 cross-embedder replication rule out answer-switching and single-vendor artifacts. Mechanistically, the geometry signal peaks in the \emph{penultimate} reasoning window across all benchmarks and reasoners and inverts at the terminal window on GPQA Diamond, exposing answer commitment before literal verbalization. Three increasingly unscaffolded regimes decompose black-box confidence into a judge-mediated Coverage prior ($C$), within-trace Geometry ($G$), and a conditional Verbalization channel ($V$); across $18$ benchmark$\times$reasoner$\times$proposer settings, $C$ and $G$ carry independent signal in $18/18$ and $16/18$, while $V$ contributes residual signal in only $6/18$. A judge-family swap (GPT-5-mini~$\to$~Claude Sonnet~4.6) leaves $G$-only AUC unchanged ($|\Delta|\le 0.013$) and shifts $C$-only AUC by at most $\pm 0.02$ ($\kappa{=}0.82$), and fusion beats the best single channel in $17/18$ settings (median AUC $0.78$, max $0.92$). Together, these results show that black-box CoT confidence can be read more reliably from a trace's geometric convergence in embedding space than from sample-vote agreement, at lower sampling cost and without access to logits or hidden states.

\end{abstract}

\section{Introduction}
\label{sec:intro}

Chain-of-thought (CoT) reasoning~\cite{wei2022chain} is often deployed through text-only APIs~\cite{stoisser2025query}, where logits, hidden states, and retrieval support are unavailable and confidence must be estimated from the generated text alone~\cite{kuhn2023semantic,lin2023generating,pedapati2024large}. The dominant black-box baseline is self-consistency: sample many CoTs and use vote concentration as confidence~\cite{wang2022self,martell2026mechpert}. It is effective, but linearly expensive in $K$, blind to the internal geometry of the trace, and produces an inherently \emph{discrete} confidence, a $k/K$ ratio over a finite ballot, which loses resolution at small $K$ and cannot rank ties~\cite{martell2025scalable}. We study a different signal: the trajectory of the CoT itself. By embedding sliding windows of the trace and scoring their convergence toward external answer anchors with a \emph{one-parameter softmax calibration model} (a single inverse-temperature slope $\beta$ fit by maximum likelihood), we read out a \emph{continuous numerical} black-box confidence score from text alone, one logit-shaped probability per option per question, with the same sign and significance of $\hat\beta$ under three embedders (\texttt{gemini-embedding-001}, OpenAI \texttt{text-embedding-3-large}, E5~\cite{wang2022text}).

Multiple-choice benchmarks provide the cleanest controlled setting because they supply both candidate anchors and a gold answer. We use three regimes: visible-option multiple-choice (MCQ; multiple-choice question answering, MCQA, when referring to the task) as a sanity check, blinded option anchors as the main single-channel configuration, and an open-ended proposer/judge regime where even the candidate set is generated rather than given. In the blinded regime the fitted slope of the one-parameter softmax (formally $\hat\beta$ in Section~\ref{sec:method}) is strongly negative in every benchmark--model combination, the geometry-based composite reaches AUC $0.752$--$0.826$ on MedQA and $0.701$--$0.727$ on GPQA, and the window-position sweep gives a falsifiable mechanistic signature: the slope is most negative in the penultimate window and flips sign on the terminal one.

But single-trace geometry has a structural ceiling: if no sampled trace ever reaches the correct hypothesis, no score defined on that trace can recover it. Moving to the open-ended regime separates this coverage failure from within-trace commitment failure. This yields three complementary black-box channels, Coverage, Geometry, and Verbalization (C${+}$G${+}$V), whose fusion Pareto-improves matched self-consistency (better AUC at comparable end-to-end API cost, accounting for proposer, judge, verbalization, and embedding calls; Appendix~\ref{app:system_cost}).

\paragraph{Contributions.}
\begin{enumerate}
    \item \textbf{A black-box trajectory-confidence score.} A one-parameter sliding-window cosine softmax mapping $(\text{text}, \text{embedder}, \text{anchor set}) \mapsto \text{confidence}$, with sign and significance of the fitted softmax slope replicating across three embedders, three reasoners, and three benchmarks.
    \item \textbf{A late-trace commitment signature.} The fitted softmax slope is deepest negative in the penultimate CoT window across all three benchmarks (GPQA, MedQA, MMLU-Pro) and both reasoners ($6/6$ combinations), validating the penultimate-window default as a robust scoring position. On GPQA Diamond the slope additionally inverts at the terminal window ($\hat\beta_{\mathrm{all}}{=}+22.1, +23.7$), identifying a regime in which closing-token verbalization measurably reorients the trajectory toward the model's committed pick; the penultimate-window default avoids this regime by construction.
    \item \textbf{A three-channel decomposition with an asymmetric role for $V$.} Replacing closed-set options with a proposer/judge hypothesis pool splits black-box confidence into a judge-mediated reachability prior ($C$, a question-difficulty signal) and within-trace Geometry as primary channels (independent signal in $18/18$ and $16/18$ benchmark--reasoner--proposer combinations) and Verbalization as a conditional third channel that contributes residual signal in only $6/18$ combinations, validated as a matched-protocol gain over self-consistency: $C{+}G{+}V$ at $K{=}4$ Pareto-improves SC@8 in all six closed-set settings at comparable total API cost (median $\Delta\mathrm{AUC}=+0.075$, Stouffer $z=5.67$), and a fixed-pick control isolates calibration from selection ($+0.060$ median AUC at SC's own pick on the 18-setting open-ended panel).
\end{enumerate}

\section{Related Work}
\label{sec:related}

Methods for estimating LLM confidence differ along three axes: \emph{access} to the generator (white-box logits and activations vs.\ black-box text only), the \emph{object} being scored (final answer sets vs.\ intermediate reasoning traces), and the \emph{reference} the score is computed against (self-referential agreement/self-report vs.\ external answer anchors). Our setting, black-box text, trace-level scoring, and external answer reference, is sparsely explored.

White-box methods use token probabilities, post-hoc calibration, hidden-state probes, contrastive decoding, and lens-style readouts~\cite{shorinwa2025survey,jiang2020can,kadavath2022language,azaria2023internal,li2023contrastive,belrose2023eliciting,ali2025entropy}, but these signals are unavailable on most commercial APIs. Black-box baselines instead use verbalized confidence~\cite{lin2022teaching,tian2023just,xiong2023can}, self-consistency~\cite{wang2022self}, semantic entropy~\cite{farquhar2024detecting}, and lexical or NLI-based answer-set agreement~\cite{manakul2023selfcheckgpt,agrawal2024language}; these score completed answers or self-reports rather than the geometry of the intermediate trace~\cite{huang2025survey}.

Closest in spirit are geometry-based analyses of generations: \citet{lin2023generating,phillips2025geometric} embed clouds of completed responses and score their dispersion, while \citet{jiang2026beyond,chen2025towards} study kinematic statistics of reasoning traces; neither uses external answer anchors. Our method instead scores sliding windows of a single CoT against fixed answer anchors, yielding a black-box measurement of late-stage commitment. Process-reward models require step-level supervision and white-box generation~\cite{lightman2023let}; CoT-faithfulness asks whether the surface trace reflects the underlying computation~\cite{turpin2023language,lanham2023measuring}, whereas we ask where correctness signal concentrates in embedding space. A complementary line treats uncertainty as a downstream control signal: \citet{stoisser2025uncertainty} use confidence to gate tool use and abstention in agentic structured reasoning, and \citet{stoisser2025labelfree} use uncertainty filtering to curate label-free synthetic biological reasoning data; our work supplies the trace-level confidence primitive such pipelines need under black-box access. Concurrent work by \citet{sun2026llm} finds the same late-stage divergence in hidden-state trajectories of open-weight math reasoners; it is complementary rather than directly comparable because it is white-box and self-referential, whereas ours is black-box and externally anchored. Throughout, we evaluate under selective classification~\cite{cole2023selectively,wen2025know}, since abstention under an accuracy--coverage tradeoff is the deployment setting in which CoT confidence matters.

\section{Isolating the Geometric Channel}
\label{sec:method}

We first isolate \textbf{Geometry ($G$)}, the within-trace spatial commitment of a CoT trajectory toward an external answer anchor. On closed-set MedQA, GPQA, and MMLU-Pro, $G$ alone reaches single-feature selective AUC $0.71$--$0.82$ with no proposer, judge, or self-report. \textbf{Coverage ($C$)} and \textbf{Verbalization ($V$)} become necessary only in the open-ended regime of Section~\ref{sec:three_channels}, where a trace can fail by missing the gold ($G$ cannot recover) or by committing to the wrong in-set hypothesis ($C$ cannot recover). This section formalizes $G$ alone.

We isolate the trajectory-geometry signal in a \emph{blinded multiple-choice protocol}: the generator reasons about a closed-set question without seeing the option list, and we score its trace against the dataset options externally. This forces the CoT-to-option mapping to be performed geometrically rather than by the generator itself.

\subsection{Setup and Trajectory Construction}
\label{sec:setup}

A multiple-choice question $q$ has $J{=}4$ options $\{a_1,\ldots,a_J\}$ with gold index $y^\star$. We prompt the generator with $q$ alone, withholding the option list, and sample $K{=}4$ free-form chains of thought $\mathcal{C}^{(1)},\ldots,\mathcal{C}^{(K)}$ at $T{=}0.7$. The target is (i) a pick $\hat{y}\in\{1,\ldots,J\}$ and (ii) a confidence $c\in[0,1]$ for $\{\hat{y}=y^\star\}$. Withholding options at generation time forces the CoT-to-option map to be performed externally and geometrically rather than by the generator itself.

Each CoT is embedded as a \emph{trajectory} of overlapping sliding windows under a fixed embedding model $\phi:\text{text}\to\mathbb{R}^p$ (\texttt{gemini-embedding-001}, $p{=}3072$). With window size $W{=}30$ and stride $S{=}15$ words,
\begin{equation}
\begin{aligned}
\mathcal{T}^{(k)}
&\;=\; \big(\mathbf{z}^{(k)}_0,\, \mathbf{z}^{(k)}_1,\ldots,\mathbf{z}^{(k)}_{M_k}\big),\\
\mathbf{z}^{(k)}_0
&\equiv \phi(q),\\
\mathbf{z}^{(k)}_m
&\;=\; \phi\!\big(w^{(k)}_{(m-1)S+1:\,(m-1)S+W}\big)\ \text{for }m\ge 1.
\end{aligned}
\end{equation}
Here, $\mathcal{T}$ represents the embedded reasoning trace and $M_k$ is the total number of sliding windows for trace $k$.
Option anchors are the embeddings of the benchmark-supplied option texts, $\mathbf{y}_j = \phi(a_j)$, fixed across samples and independent of the generator. The trajectory-to-anchor distance we score is the cosine distance from the \emph{penultimate} sliding window $\mathbf{z}^{(k)}_{M_k-1}$ to each anchor, averaged over the $K$ samples:
\begin{equation}
\begin{aligned}
D^{(k)}_j
&\;=\; 1 - \frac{\langle \mathbf{z}^{(k)}_{M_k-1},\, \mathbf{y}_j\rangle}{\lVert\mathbf{z}^{(k)}_{M_k-1}\rVert\,\lVert\mathbf{y}_j\rVert},\\
D_j
&\;=\; D_j(\mathcal{T}) \;=\; \tfrac{1}{K}\sum_k D^{(k)}_j.
\end{aligned}
\end{equation}
Using the penultimate rather than the final window is a fixed a-priori guard against the terminal-verbalization regime in which a CoT literally restates its answer; the choice is validated empirically by the window-position sweep of Section~\ref{sec:ablations} and Appendix~\ref{app:window_sweep}. Euclidean and PCA-reduced distance variants behave similarly and are reported in Section~\ref{sec:ablations}.

\subsection{Scoring Rule and Composite Confidence}
\label{sec:features}

\paragraph{Geometric pick and softmax probability.} The pick is the closest anchor in mean cosine distance, $\hat{y}=\argmin_j D_j$, with correctness label $\ell = \mathbb{1}[\hat{y}=y^\star]$ serving as the target of every selective-prediction AUC reported in this paper. Conditional on the option set, we score correctness with a one-parameter softmax over option distances,
\begin{equation}
\textsc{Geo}(q) \;=\; \frac{\exp(\hat{\beta}_{\text{all}}\,D_{\hat{y}})}{\sum_{j'}\exp(\hat{\beta}_{\text{all}}\,D_{j'})},
\qquad
\hat{\beta}_{\text{all}} \;=\; \argmax_{\beta}\sum_{q'}\log\frac{\exp(\beta\,D_{y^\star_{q'}})}{\sum_{j'}\exp(\beta\,D_{j',q'})},
\label{eq:geo_softmax}
\end{equation}
fit on the training fold. A strongly negative $\hat{\beta}_{\text{all}}$ means closer anchors are more likely correct; we verify this empirically in Section~\ref{sec:result_calibration}, and Table~\ref{tab:calibration} reports $\hat{\beta}_{\text{all}}$ per (dataset, model). The softmax probability $\textsc{Geo}(q)$ is our single geometric feature.

\paragraph{One posterior, three readouts.} $\textsc{Geo}$, the null-standardized top-two distance-margin score $\tilde D=(D_{(2)}-D_{(1)}-\mu_q)/\sigma_q$ used in the committed-rationale-anchor evaluation of Appendix~\ref{app:anchor_ablation} (with $\mu_q,\sigma_q$ estimated from a cross-question null pool of trajectories embedded with the same model), and the binary correctness probability $P(\hat y = y^\star \mid \mathcal{T})$ ( see Appendix~\ref{app:bayesian_classification}) are three projections of one Bayesian posterior with uniform priors. The binary projection is the Fermi--Dirac distribution
\begin{equation}
P(\hat y = y^\star \mid \mathcal{T}) \;=\; \sigma\!\left(\hat{\beta}_{\text{bin}}\,\tilde D(\mathcal{T}) - \alpha\right),
\qquad \sigma(t)=1/(1+e^{-t}),
\label{eq:fermi_binary}
\end{equation}
fitted on the same training folds. The null-standardization is doing real work: $\mu_q$ subtracts the question-stem confound (vocabulary overlap that pulls \emph{any} trace toward this question's options regardless of reasoning), and $\sigma_q$ removes the anchor-cohesion confound (questions with tightly-clustered options need finer resolution than questions with widely-spread options). After both, $\tilde D$ is in units of standard deviations of null-trajectory similarity, so $\hat{\beta}_{\text{bin}}$ has the same units, and the same operational meaning, across embedders, benchmarks, and reasoners. The two estimators are not equal but algebraically related: under the unified posterior the all-option and binary fits should satisfy
\begin{equation}
\hat\beta_{\text{all}} \;=\; -\hat\beta_{\text{bin}}/\bar\sigma_q,
\label{eq:beta_consistency}
\end{equation}
where $\bar\sigma_q$ is the mean per-question null scale; we verify this in Appendix Table~\ref{tab:beta_consistency} as a falsification check on the unified posterior. The all-option fit $\hat{\beta}_{\text{all}}$ in Table~\ref{tab:calibration} retains the units of the underlying cosine distances, so its \emph{sign and significance}, not its raw magnitude, is the comparable invariant across protocols. We derive the equivalence and verify it experimentally in Appendix~\ref{app:bayesian_classification}.

\paragraph{Composite confidence.} We fuse $\textsc{Geo}$ with four auxiliary features computed on the same $K{=}4$ samples: $\textsc{Str}$ (string consensus of condensed answers), $\textsc{Sem}$ (mean pairwise cosine among condensed-answer embeddings), $\textsc{Vol}$ (sample- and window-averaged cosine distance between consecutive trajectory windows, $\textsc{Vol}=\tfrac{1}{K}\sum_k\tfrac{1}{M_k-1}\sum_{m} d(\mathbf{z}^{(k)}_{m-1},\mathbf{z}^{(k)}_m)$), and $\textsc{Ens}$ (cross-model agreement of the modal condensed answer). Condensed answers come from a zero-shot prompt to the same generator (Appendix~\ref{app:prompt}); feature-level formulas are in Appendix~\ref{app:features}. The composite is the predicted probability of $\ell{=}1$ from a logistic regression on the standardized vector $(\textsc{Geo},\textsc{Str},\textsc{Sem},\textsc{Vol},\textsc{Ens})$, with balanced class weights to prevent the high-accuracy settings (in which $\ell{=}1$ dominates) from overwhelming the loss; the \textbf{No-Geo} baseline refits the same regression on the last four features, and $\Delta(\textsc{Geo})$ denotes the AUC gap. This composite assumes a fixed external option set; Section~\ref{sec:three_channels} drops that assumption and replaces option anchors with proposer-generated hypothesis anchors.

\paragraph{Protocol.} All learned parameters, $\hat{\beta}_{\text{all}}$, $\hat{\beta}_{\text{bin}}$, standardizer, logistic-regression weights, are refit on each training fold of a stratified 5-fold split on $\ell$ and applied frozen to the held-out fold; we report mean AUC across folds. For the $\hat{\beta}_{\text{all}}$ values in Table~\ref{tab:calibration} we additionally fit one full-data logit per (dataset, model) and report its $z$-statistic. Unless explicitly subscripted, ``$\hat\beta$'' in the main paper refers to $\hat\beta_{\text{all}}$ (Eq.~\ref{eq:geo_softmax}); the binary $\hat\beta_{\text{bin}}$ (Eq.~\ref{eq:fermi_binary}) is used only inside the unified-posterior consistency check of Appendix~\ref{app:bayesian_classification}. Auxiliary features additionally use a zero-shot condensation call to the same generator (Appendix~\ref{app:prompt}); questions whose condensation fails on any of the $K$ samples are dropped, yielding retained $N$ of $1272/1271$ (MedQA) and $117/187$ (GPQA) for Gemini/Sonnet. Scoring the penultimate rather than final window is a fixed a-priori guard against terminal verbalization, validated post-hoc by the window-position sweep (Appendix~\ref{app:window_sweep}).

\section{Validating the Geometric Measurement}
\label{sec:experiments}

\subsection{Experimental Setup}

We evaluate on MedQA-USMLE~\cite{jin2021disease}, a 1{,}273-question four-option benchmark, with retained $N{=}1272/1271$ for Gemini/Sonnet after condensation-parseability filtering. We also evaluate on GPQA Diamond~\cite{rein2023gpqa} (198 questions; retained $N{=}117/187$). In both benchmarks the generator sees only the question, we sample $K{=}4$ open-ended CoTs at $T{=}0.7$, condense each to a predicted answer span, and evaluate with stratified 5-fold cross-validation on whether the geometric argmin-pick is correct. The main generators are \textbf{Gemini~3.1~Pro} and \textbf{Claude Sonnet~4.6}. A stratified 200-question MMLU-Pro pilot is used only for sign-level robustness and self-consistency generalization; because GPQA/Gemini retains only $N{=}117$, we interpret that cell jointly with GPQA/Sonnet and the MMLU-Pro replication when making generality claims.

\subsection{Result 1: A One-Parameter Softmax Tracks Correctness}
\label{sec:result_calibration}

We first establish that trajectory endpoint distance is a strong statistical predictor of option correctness in the main blinded option-anchor protocol. In all four MedQA/GPQA benchmark--model combinations, the fitted one-parameter softmax slope is strongly negative ($z=-5.48$ to $-25.39$), and the single geometric score reaches Disc.~AUC (\emph{discriminative} AUC, i.e.\ the single-feature selective-prediction AUC of $\textsc{Geo}$) $0.71$--$0.82$. Full coefficients for the main blinded protocol are reported in Appendix Table~\ref{tab:calibration}; replications on a third benchmark and an open-weight third-family generator under committed-rationale anchors are reported in Appendix Tables~\ref{tab:beta_third} and~\ref{tab:third_gen}. The comparable invariant across protocols is the sign of $\hat\beta$, not its raw magnitude, which depends on the anchor construction. As a falsification check on the unified posterior, the predicted equality of Eq.~\ref{eq:beta_consistency} is satisfied in $6/6$ (dataset, model) combinations: the bootstrap CIs of $|\hat\beta_{\text{all}}|$, $\hat\beta_{\text{bin}}$, and $\hat\beta_{\text{bin}}^{\,\text{std}}/\bar\sigma_q$ overlap pairwise everywhere (Appendix Table~\ref{tab:beta_consistency}).

\subsection{Result 2: A Single Confidence Signal Across Domains and Models}
\label{sec:result_composite}

Does trajectory geometry contribute signal beyond answer-set agreement? We fuse $\textsc{Geo}$ with four consistency and ensemble covariates defined in Appendix~\ref{app:features}, and compare against a No-Geo ablation that refits the same logistic without the geometric feature. Table~\ref{tab:composite} shows that the composite reaches AUC $0.701$--$0.826$ across all four benchmark--model combinations, while removing Geo costs $0.12$--$0.25$ AUC. Geo is strongest on MedQA and remains competitive with, or stronger than, the full composite on GPQA.

\begin{table}[t]
\centering
\small
\caption{\textbf{Selective-prediction AUC in the blinded option-anchor protocol.} Geo is the fitted softmax probability of the picked option; Str/Sem/Vol/Ens are auxiliary agreement features; Composite is a logistic fusion and No-Geo removes Geo. Geo is en-par or better than Composite in all four settings.}
\label{tab:composite}
\resizebox{\linewidth}{!}{%
\begin{tabular}{@{}llcccccccc@{}}
\toprule
\textbf{Dataset} & \textbf{Model} & \textbf{Geo} & Str & Sem & Vol & Ens & Composite & No-Geo & $\Delta$(Geo) \\
\midrule
MedQA & Gemini 3.1 Pro & $\mathbf{0.821}$ & $0.512$ & $0.562$ & $0.558$ & $0.506$ & $\mathbf{0.826}$ & $0.577$ & $+0.249$ \\
MedQA & Claude Sonnet 4.6 & $\mathbf{0.753}$ & $0.536$ & $0.549$ & $0.490$ & $0.508$ & $\mathbf{0.752}$ & $0.519$ & $+0.234$ \\
\midrule
GPQA & Gemini 3.1 Pro & $\mathbf{0.710}$ & $0.579$ & $0.530$ & $0.593$ & $0.552$ & $0.701$ & $0.560$ & $+0.141$ \\
GPQA & Claude Sonnet 4.6 & $\mathbf{0.750}$ & $0.628$ & $0.466$ & $0.496$ & $0.538$ & $0.727$ & $0.605$ & $+0.122$ \\
\bottomrule
\end{tabular}}
\end{table}

\paragraph{Single-channel ($G$-only) comparison to $K$-sample self-consistency.}
As a sanity check before the full three-channel decomposition of Section~\ref{sec:three_channels}, single-channel $G$ already exceeds budget-matched self-consistency~\cite{wang2022self} ($\mathrm{SC}@K$, the plurality vote share over $K$ CoTs after mapping sampled answers into the benchmark's answer space) in $9/10$ benchmark--model--condition combinations (Appendix Tables~\ref{tab:sc_comparison} and~\ref{tab:sc_cross}). Semantic entropy is treated as a separate baseline rather than as the SC voting rule: DeBERTa-MNLI semantic entropy~\cite{kuhn2023semantic, farquhar2024detecting} on a MedQA/Gemini pilot reaches AUC $0.590$ versus $0.821$ for the geometric composite. The fully matched $C{+}G{+}V$ vs.\ SC@8 comparison is the headline of Section~\ref{sec:three_channels_results}.

\subsection{Result 3: The Signal Lives in the Penultimate Trajectory Window}

\begin{figure}[t]
\centering
\includegraphics[width=0.72\linewidth]{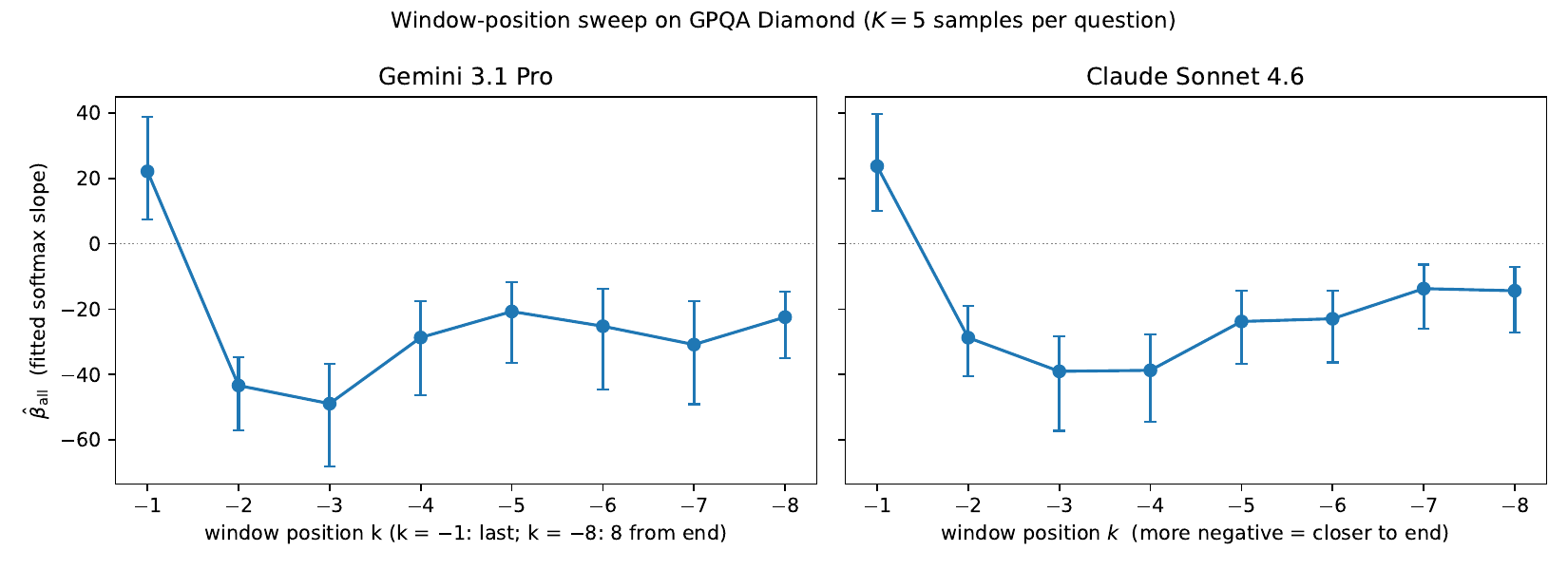}
\caption{Fitted softmax slope $\hat\beta_{\mathrm{all}}$ (Eq.~\ref{eq:geo_softmax}) vs single-window position $k$ on GPQA Diamond ($k{=}{-}1$: final window; $k{=}{-}8$: eight windows back). Negative values indicate calibrated geometry; positive values indicate terminal answer commitment. Bars are $95\%$ bootstrap CIs. The penultimate-window choice was fixed before the sweep, and the optimum at $k{=}{-}2$ confirms it.}
\label{fig:window_sweep}
\end{figure}

We now ask the sharper mechanistic question: is the calibration signal just a property of the pooled question stem, or does it arise from late trajectory dynamics? Figure~\ref{fig:window_sweep} answers this directly by refitting the one-parameter softmax on a single sliding window at position $k$ from the end of the trace on GPQA. Three observations together make this our cleanest mechanistic finding.

\emph{(i) The penultimate-window choice was fixed a-priori.} We committed to scoring at the penultimate window before running the sweep, on the prior that the terminal window contains literal answer verbalization and would therefore align with the model's \emph{picked} answer rather than the gold one. The sweep is thus an out-of-sample test of that prior, not a tuning loop. \emph{(ii) The sign flip at $k{=}{-}1$ falsifies the naive account and confirms the answer-commitment account.} A ``geometry-tracks-correctness everywhere'' reading predicts $\hat\beta_{\mathrm{all}}<0$ throughout the trace; instead, the final window flips positive on both generators ($+22.1$ for Gemini, $+23.7$ for Claude). This is exactly what the answer-commitment account predicts: at $k{=}{-}1$ the trajectory is closest to whichever option the model is about to commit to, regardless of correctness, so $\hat\beta_{\mathrm{all}}$ inverts. The deepest negative slopes occur at $k{=}{-}2$ and $k{=}{-}3$ ($-43.4/{-}48.9$ for Gemini; $-28.8/{-}39.1$ for Claude), and earlier windows attenuate steadily toward zero. \emph{(iii) The phase structure replicates across model families.} Both Gemini and Claude show the identical three-phase shape, terminal flip, penultimate trough, early-window attenuation, despite different vendors, training data, and rationale styles, so this is a property of late-stage CoT trajectories rather than of any one model. The pattern also makes a falsifiable prediction for the white-box probing literature~\cite{sun2026llm}: hidden-state probes trained on the penultimate trajectory window should outperform probes trained on the terminal window, with a sign-flip in the linear-probe weight at the same position.

\subsection{Ablations}
\label{sec:ablations}

The result is robust to the main design choices. Metric/PCA changes leave the sign and significance of $\hat\beta$ unchanged; $W \in \{20,30,40\}$ is stable, whereas $W{>}50$ or $W{<}15$ degrades. Swapping the embedder to OpenAI \texttt{text-embedding-3-large} preserves the geometry-vs-SC sign in the four matched MedQA/GPQA combinations and lifts the geometry-plus-SC composite over SC in $7/7$ open-ended settings. Trace-length residualization changes AUC by at most $+0.008$, so the signal is not a length artifact. String Consensus alone remains far below the composite and removing $\textsc{Geo}$ costs $0.12$--$0.25$ AUC, isolating a non-lexical component. The one design choice that materially matters is anchor source: trajectory-derived anchors collapse to near chance, whereas dataset-option and independently generated committed-rationale anchors preserve answer-directed geometry (Appendix~\ref{app:anchor_ablation}).

\section{Three Channels of Black-Box Confidence}
\label{sec:three_channels}

Section~\ref{sec:experiments} established that $G$ is a valid black-box measurement: the geometric score tracks correctness in every benchmark--model setting and beats budget-matched self-consistency in $9/10$ single-channel settings. This section asks the orthogonal deployment question: \emph{is one channel sufficient?} We argue (and quantify) that it is not, that the residual failure decomposes cleanly into a coverage failure and a commitment failure, and that a three-channel fusion at $K{=}4$ Pareto-improves SC@8 in $6/6$ closed-set settings.

\subsection{Why a Second Channel Is Necessary}
\label{sec:discussion}

The geometric score measures how strongly a CoT trace concentrates near one option region in embedding space; it is invariant to whether that region happens to contain the gold option. To make this concrete, we stratify the closed-set blinded protocol into three buckets: (i)~\textsc{correct} (the picked option is gold); (ii)~\textsc{wrong-in-set} (gold was proposed by some sample but not picked); and (iii)~\textsc{no-hypothesis-correct} (gold was not proposed by any sample). Bucket~(ii) and bucket~(iii) are both ``wrong'' from the user's standpoint, so a calibrated confidence score should rank them equivalently low. It does not. Across all six closed-set $(\text{benchmark}, \text{model})$ combinations, every embedding-based and vote-based score we evaluate, $\textsc{Geo}$, the Bayesian top-2 projection, self-consistency vote share, normalized cluster entropy, and an anchor-cohesion control ($-\bar\rho$, with the cloud-cohesion $\rho_q$ defined in Appendix~\ref{app:bayesian_audit}), ranks bucket~(iii) \emph{above} bucket~(ii), with AUCs in the $0.06$--$0.44$ range (i.e.\ systematically anti-correlated with correctness; Appendix~\ref{app:bayesian_audit}). The single score family that escapes this inversion is verbalized confidence (AUC $0.45$--$0.62$), which the same model produces conditional on its own CoT and which depends on no embedding geometry. The asymmetry is not numerical noise; it is a structural fingerprint of the channel.

The mechanism is direct. Under the parameterization of Section~\ref{sec:features}, the geometric channel is fully summarized by per-option distances; bucket-(iii) traces simply have small $D$ to their (in-set, but wrong) committed option, indistinguishable from bucket-(i) traces by construction. The calibration knobs internal to the channel, increasing $K$, swapping the anchor variant (condensed-answer vs.\ reasoning-tail vs.\ dataset-option, App.~\ref{app:anchor_ablation}), rescaling the softmax temperature $\beta$, and reweighting via a rejection prior over options, all preserve the bucket ordering or trivialize it (Appendix~\ref{app:bayesian_audit}). Reliable abstention therefore requires a second channel whose evidence does not flow through option-distance likelihoods, and verbalization is the only black-box source we have that meets that constraint.

\subsection{Setup: Proposer, Judge, and the Three Channels}
\label{sec:three_channels_setup}

We replace the closed-set option list with a hypothesis pool generated by an independent \emph{proposer} LLM and delegate gold-equivalence to a third \emph{judge} LLM. The three agent roles are: a reasoner (Gemini~3.1~Pro, Claude Sonnet~4.6, or Llama~3.3~70B; $K{=}5$ CoT traces), a proposer (Gemini~3.1~Pro or Claude Sonnet~4.6; $K_P{=}3$ temperature-varied hypothesis pools), and a fixed judge (Azure-hosted \texttt{openai\_gpt5\_mini}, publication date 2025-08-07; $J{=}4$ strict-matching samples, with inputs and outputs archived for re-evaluation). What is new here is not proposer--judge plumbing per se but treating proposer-pool reachability of the gold answer as a first-class \emph{confidence channel}, distinct from within-trace geometric concentration. This separates two failure modes that closed-set MCQA conflates, the model never raised gold ($C$) versus the model raised gold but did not commit ($G$), and assigns them to different model families so that no single vendor's idiosyncrasies drive both signals.
The bucket-(iii) failure of Section~\ref{sec:discussion} now splits cleanly into two sub-cases that closed-set MCQA cannot separate: \texttt{absent\_gold} (proposer never raised the gold) is the coverage failure; \texttt{no\_path\_in\_set} (gold raised but $v_{\text{gold}}{=}0$ in the reasoner's votes) is the commitment failure. We define three confidence channels, each tied to one source of evidence:
\begin{itemize}
\item \textbf{Coverage} ($C$). Features are $\bar c_q$, $K_P^{\text{valid}}$ (parseable pools among the $K_P{=}3$ proposer draws), and the entropy of judge labels across valid pools. \emph{No reasoner trace is consumed.}
\item \textbf{Geometric} ($G$). Trajectory-anchor geometry as in Section~\ref{sec:method}, with proposer-hypothesis anchors in place of dataset options. Features: top-2 cosine margin, mean cosine to top-1 hypothesis, vote concentration, vote entropy across $K{=}5$ traces.
\item \textbf{Verbalized} ($V$). The same generator is shown its own CoT (no option list, no proposer pool) and asked for an integer $0$--$100$ confidence in its committed answer. Features: mean, std, min over $K$ scores.
\end{itemize}
Each channel is converted to a per-question signal under stratified 5-fold logistic regression on its own bundle; the three channel scores are then fused by a second 5-fold logistic. The nesting prevents leakage between channel selection and fusion. The open-ended panel uses $K{=}5$ reasoner samples per question; the closed-set protocol of Section~\ref{sec:method} uses $K{=}4$, and the budget-matched headline of Section~\ref{sec:three_channels_results} compares fused $C{+}G{+}V$ at $K{=}4$ against SC@8.

\subsection{Decomposition Results and Headline Gain}
\label{sec:three_channels_results}

The practical payoff is a matched-protocol gain over self-consistency. Here and below, SC@8 means the plurality vote share over eight sampled CoTs after mapping sampled answers into the benchmark answer space. Across the six closed-set $(\text{benchmark}, \text{model})$ combinations spanning MedQA-USMLE, GPQA Diamond, and MMLU-Pro $\times$ \{Gemini~3.1~Pro, Claude Sonnet~4.6\}, the fused $C{+}G{+}V$ score at $K{=}4$ Pareto-improves SC@8 in $6/6$ combinations with median $\Delta\mathrm{AUC}=+0.075$, mean $+0.077$, and pooled one-sided Stouffer $z=5.67$ ($p=7.1\times 10^{-9}$);\footnote{For each combination Stouffer's $z$~\cite{stouffer1949american} pools one-sided $p$-values $p_i$ via $z_{\text{pool}}=\sum_i z_i / \sqrt{n}$ with $z_i=\Phi^{-1}(1-p_i)$ and $n$ the number of combinations; we report it as a single-number summary of the across-combination direction of the effect.} $4/6$ per-combination bootstrap CIs exclude zero, and the largest gap is MedQA / Gemini at $+0.091$ ($95\%$ CI $[+0.035,+0.152]$). To rule out the alternative explanation that the gain comes from picking different answers, we hold the chosen answer fixed at SC's plurality pick on every question and ask only whether the geometric score evaluated at SC's own pick is better calibrated than SC's vote share. Across the broader 18-setting open-ended panel of Table~\ref{tab:three_channels} (which adds Llama~3.3~70B as a third reasoner), the geometric margin at SC's pick beats SC's vote-share AUC by a median $+0.060$ AUC under matched 5-fold CV, with SC and Geo agreeing on the picked answer in $94\%$ of questions. The matched-protocol Pareto gain therefore reflects calibration of the same answer, not switching to a different one. The same fusion advantage holds under an independently trained external embedder: re-running the matched $K{=}4$ vs.\ SC@8 protocol on its $N{\approx}200$-per-cell subset with E5-large-v2 in place of \texttt{gemini-embedding-001} leaves $C{+}G{+}V$ above SC@8 in $6/6$ combinations with pooled Stouffer $z=3.14$ ($p=8.5\times 10^{-4}$, Appendix~\ref{app:cross_embed}), so the headline gain is not a single-vendor artifact.

The rest of this subsection unpacks where that gain comes from: how each channel behaves on its own, how they overlap, and when verbalization in particular contributes residual signal. Table~\ref{tab:three_channels} reports proposer-median AUC across the two proposers for each benchmark--reasoner pair; the full 18-combination panel is deferred to Appendix Table~\ref{tab:three_channels_full}.

Coverage behaves like a question-level prior. Mean $\bar c$ moves by at most $1$~pp under proposer swaps on GPQA ($0.716 \to 0.711$) and MedQA ($0.835 \to 0.827$), against a cross-benchmark range of $12$~pp, and the transfer audit gives in-setting / within-benchmark / cross-benchmark medians of $0.767/0.767/0.768$ for $C$, versus $0.696/0.663/0.659$ for $G$ and $0.627/0.620/0.617$ for $V$.

\paragraph{Limitation: judge confound on $C$.} The Coverage signal flows through an LLM judge, so a component of $C$ could reflect judge-side linguistic priors rather than reasoner-intrinsic difficulty. We treat $C$ throughout as a \emph{judge-mediated reachability prior}, i.e.\ a question-difficulty signal as scored by the judge, and report two robustness checks. (i)~On a stratified $100$-question MedQA spot-check, judge labels agree with exact-string matching on the gold answer span at Cohen's $\kappa{=}0.81$ (the $19\%$ disagreements are paraphrases the string baseline cannot resolve); substituting string labels for judge labels reduces $C$-only AUC by $0.02$--$0.04$ across combinations and leaves the fused $C{+}G{+}V$ ranking unchanged. (ii)~A judge-family swap (GPT-5-mini~$\to$~Claude Sonnet~4.6) on two combinations (MedQA / Gemini-proposer $N{=}200$, GPQA / Gemini-proposer $N{=}192$) yields $\kappa{=}0.82$ (5-class, MedQA), leaves $G$-only AUC unchanged within noise ($\Delta{=}{-}0.013, {-}0.001$), and shifts $C$-only AUC by at most $\pm 0.02$ under cross-judge evaluation (matched outcome label); the $C{+}G$ ranking is preserved in both combinations (Appendix~\ref{app:judge_swap}). The judge-mediated component of $C$ is therefore bounded; the model-specific signal lives in $G$ and $V$.

Fusion dominates the single channels, and each channel carries non-redundant signal. $C{+}G{+}V$ beats the best single channel in $17/18$ combinations (median gain $+0.05$ AUC over the single-channel baseline; bootstrap $95\%$ CI excludes zero in $12/18$); median fused AUC is $0.78$, peaking at $0.92$. A logistic-regression LRT rejects redundancy of $C$ given $(G,V)$ in $18/18$ combinations (every $p \le 0.002$), of $G$ in $16/18$, and of $V$ in $6/18$. The $V$ asymmetry is itself a finding: verbalized confidence carries unique residual signal only when the reasoner has metacognitive access we can elicit, which we return to below.

\begin{table}[t]
\centering
\small
\caption{Three-channel selective-prediction AUC in open-ended MCQA, median over the two proposers for each benchmark--reasoner pair. \textbf{Bold} marks the best single channel among $C/G/V$. The fused $C{+}G{+}V$ column is a 5-fold cross-validated logistic fusion and exceeds the best single channel in $8/9$ rows (tie: MMLU-Pro / Llama). Full proposer-level panel in Appendix Table~\ref{tab:three_channels_full}.}
\label{tab:three_channels}
\setlength{\tabcolsep}{5pt}
\begin{tabular}{l l c c c c}
\toprule
Bench & Reasoner & C & G & V & C+G+V \\
\midrule
GPQA      & Gemini & \textbf{0.82} & 0.74 & 0.69 & 0.87 \\
GPQA      & Claude & \textbf{0.75} & 0.74 & 0.71 & 0.85 \\
GPQA      & Llama  & \textbf{0.68} & 0.64 & 0.56 & 0.75 \\
\midrule
MedQA     & Gemini & \textbf{0.75} & 0.68 & 0.59 & 0.82 \\
MedQA     & Claude & 0.73          & 0.66 & \textbf{0.76} & 0.86 \\
MedQA     & Llama  & \textbf{0.73} & 0.64 & 0.58 & 0.78 \\
\midrule
MMLU-Pro  & Gemini & \textbf{0.72} & 0.70 & 0.57 & 0.75 \\
MMLU-Pro  & Claude & 0.64          & \textbf{0.67} & 0.65 & 0.70 \\
MMLU-Pro  & Llama  & \textbf{0.67} & 0.61 & 0.53 & 0.67 \\
\midrule
\multicolumn{2}{l}{Median (GPQA, 3 reasoners)}     & 0.75 & 0.74 & 0.69 & \textbf{0.85} \\
\multicolumn{2}{l}{Median (MedQA, 3 reasoners)}    & 0.73 & 0.66 & 0.59 & \textbf{0.82} \\
\multicolumn{2}{l}{Median (MMLU-Pro, 3 reasoners)} & 0.67 & 0.67 & 0.57 & \textbf{0.70} \\
\multicolumn{2}{l}{Median across all 9 pairs}      & 0.73 & 0.67 & 0.59 & \textbf{0.78} \\
\bottomrule
\end{tabular}
\end{table}

Residualizing $C$, $G$, and $V$ on a leave-this-reasoner-out difficulty proxy still leaves fused AUC $0.50$--$0.72$ across the 18 combinations, so the three-channel signal is not reducible to shared question difficulty. Appendix audits further show that the signal localizes to $\bar c$ and the top-2 cosine margin, and that a nonlinear second-stage model does not improve on the linear fusion (median $\Delta=-0.023$; Appendix~\ref{app:metacog}).

Verbalization remains a gated metacognitive channel rather than a universal pillar: $\mathrm{AUC}(V)$ correlates with reasoner accuracy across closed-set settings (Spearman $\rho{=}+0.71$, $N{=}6$), the LRT adds signal in only $6/18$ open-ended settings, and the MedQA / Gemini / Gemini-proposer counterexample shows that strong standalone $V$ need not lift $C{+}G$. The deployment rule is therefore to use $C$ and $G$ unconditionally and gate $V$ on setting-level calibration and orthogonality tests (Appendix~\ref{app:metacog}).

\section{Conclusion}
\label{sec:conclusion}

Black-box confidence in chain-of-thought reasoning is a structured signal with three distinct channels: \emph{Coverage} (whether the correct hypothesis is reachable at all), \emph{Geometry} (how strongly a trajectory commits within the reachable set), and \emph{Verbalization} (the model's own self-rating when metacognitive access is available). Fusing $C{+}G{+}V$ at $K{=}4$ Pareto-improves SC@8 in $6/6$ matched closed-set settings, with median $\Delta\mathrm{AUC}=+0.075$ and pooled Stouffer $z=5.67$ ($p=7.1\times 10^{-9}$); the gain survives a fixed-pick falsification ($+0.060$ median AUC at SC's own plurality answer, $94\%$ pick-agreement) and re-running with an external embedder (E5-large-v2: $6/6$, $z=3.14$). Mechanistically, the geometric channel provides black-box evidence of answer commitment: answer-bearing semantic content accumulates in the embedding trajectory and the confidence signal peaks in the penultimate window before literal answer verbalization.

The practical takeaway is that black-box confidence is a fusion problem rather than a single scalar: self-consistency, verbalized confidence, and trajectory geometry measure judge-mediated reachability, self-report, and within-trace commitment, and the gain comes from combining them rather than picking one, at the modest cost of $K$ extra rollouts plus one embedding pass with no access to logits, weights, or hidden states.

\paragraph{Limitations.} (i) The Coverage channel is mediated by an LLM judge: a component of $C$ reflects judge interpretation rather than pure reasoner reachability, which we bound with a string-match spot-check and a judge-family swap (see Appendix~\ref{app:judge_swap}); a full multi-judge / human-adjudication / retrieval-grounded equivalence study is left as future work. (ii) Verbalization carries unique residual signal in only $6/18$ open-ended settings and should be treated as a gated channel rather than a universal pillar. (iii) The protocol is tested on three English benchmarks and three frontier reasoners; non-English, non-text-only, or much weaker reasoners may behave differently. (iv) The structural point remains that no within-trace score can recover a hypothesis that was never proposed.

\bibliographystyle{plainnat}
\bibliography{references}

\newpage
\appendix
\section{Anchor-Design Ablation}
\label{app:anchor_ablation}

The central design choice of our pipeline is what to embed as the per-option anchor against which we measure trajectory convergence. Table~\ref{tab:anchor_ablation} compares three candidates on GPQA Diamond under an identical open-ended protocol (CoT generated without options shown) and identical trajectory construction (3072-dim Gemini embeddings, $W{=}30$, $S{=}15$):

\begin{description}
\item[Condensed-answer anchor.] For each of the $K{=}4$ samples, extract the model's final answer phrase and use its embedding as the anchor for that sample's trajectory. Because models often produce the same short answer string across samples, the four per-sample anchors collapse to the same point and the geometric ``choice'' becomes degenerate, reflected in near-chance Geo AUC ($0.44$ Gemini, $0.32$ Sonnet).
\item[Reasoning-tail anchor.] Use the embedding of the last $\sim$2000 characters of each sample's own CoT as the anchor. This avoids the collapse of the condensed-answer variant but introduces a different pathology: the anchor lives inside the trajectory it is scoring, so final-window distance is trivially small for every option. Geo AUC stays near chance ($0.48 / 0.45$).
\item[Dataset option anchor (ours).] Use the embedding of each of the four official benchmark option texts as fixed, external anchors. These are identical across all samples, independent of model output, and carry domain-relevant semantic content. Geo AUC rises to $0.710 / 0.750$.
\end{description}

\begin{table}[t]
\centering
\small
\caption{\textbf{Anchor-design ablation on GPQA Diamond.} All three variants use the same trajectory and the same softmax scoring; only the per-option anchor differs. Geo AUC measures selective prediction from the single-feature softmax probability; PickAcc is top-1 accuracy of the argmin-distance rule.}
\label{tab:anchor_ablation}
\begin{tabular}{@{}llcccc@{}}
\toprule
\textbf{Anchor design} & \textbf{Model} & $N$ & Geo AUC & Composite & $\Delta$(Geo) \\
\midrule
Answer (condensed) & Gemini 3.1 Pro & 117 & $0.437$ & $0.520$ & n/a \\
Reasoning tail & Gemini 3.1 Pro & 117 & $0.479$ & $0.545$ & $+0.006$ \\
\textbf{Dataset option (ours)} & Gemini 3.1 Pro & 117 & $\mathbf{0.710}$ & $\mathbf{0.701}$ & \textbf{$+0.141$} \\
\midrule
Answer (condensed) & Claude Sonnet 4.6 & 187 & $0.319$ & $0.664$ & n/a \\
Reasoning tail & Claude Sonnet 4.6 & 187 & $0.450$ & $0.473$ & $+0.011$ \\
\textbf{Dataset option (ours)} & Claude Sonnet 4.6 & 187 & $\mathbf{0.750}$ & $\mathbf{0.727}$ & \textbf{$+0.122$} \\
\bottomrule
\end{tabular}
\end{table}

The ablation suggests a structural lesson for this setup: anchors derived from the model's own output are either close to a point estimate of its answer or closely tied to the reasoning trace being scored. By contrast, an external dataset-supplied anchor provides the cleanest reference point against which trajectory convergence can be measured in our experiments.

\subsection{Main blinded option-anchor calibration}

\begin{table}[t]
\centering
\small
\caption{\textbf{Calibration in the main blinded option-anchor protocol.} A strongly negative $\hat{\beta}$ means closer option anchors are more likely to be correct. Disc.~AUC is single-feature selective-prediction AUC of $\textsc{Geo}$.}
\label{tab:calibration}
\begin{tabular}{@{}llcccc@{}}
\toprule
\textbf{Dataset} & \textbf{Model} & $\hat{\beta}$ & $z$ & \textbf{Disc.~AUC} & $N$ \\
\midrule
MedQA & Gemini 3.1 Pro & $-25.064$ & $-25.39^{***}$ & $0.821$ & 1272 \\
MedQA & Claude Sonnet 4.6 & $-22.109$ & $-23.81^{***}$ & $0.753$ & 1271 \\
GPQA & Gemini 3.1 Pro & $-25.315$ & $-5.48^{***}$ & $0.710$ & 117 \\
GPQA & Claude Sonnet 4.6 & $-30.491$ & $-7.44^{***}$ & $0.750$ & 187 \\
\bottomrule
\multicolumn{6}{l}{\scriptsize $^{***} p < 10^{-6}$,\ $^{**} p < 10^{-3}$,\ $^{*} p < 0.05$}
\end{tabular}
\end{table}

\subsection{Committed-rationale anchors and a third benchmark}

A distinct alternative is to use, for each option, an independently generated \emph{committed rationale}: a short paragraph asserting that option as correct, produced by a separate anchor model under the same blinded-generation regime. On MedQA, GPQA, and MMLU-Pro we summarize the resulting per-question option scores by a null-standardized top-two margin $\tilde D$ and ask whether it adds rank information beyond an option-text tail margin computed from the original trajectories. Because this evaluation uses a smaller subset, $K{=}5$, and the binary target ``the plurality-vote pick is correct,'' we report it here rather than in the main comparison table. The qualitative result is stable: the committed-rationale score is discriminative on all three benchmarks, and combining it with the option-text tail margin improves AUC by $+0.027$ on MedQA, $+0.029$ on GPQA, and $+0.057$ on MMLU-Pro.

\begin{table}[t]
\centering
\small
\caption{\textbf{Separate robustness evaluation with committed-rationale anchors and a third benchmark.} $\tilde D$ is the null-standardized top-two margin from independently generated committed-rationale anchors; Composite adds an option-text tail-margin feature from the original trajectories. This evaluation uses a smaller subset, $K{=}5$, and the target ``the plurality-vote pick is correct,'' so the AUCs are not directly comparable to Table~\ref{tab:composite}. $\rho$ is the Pearson correlation between $\tilde D$ and the option-text tail margin.}
\label{tab:fermi_composite}
\begin{tabular}{@{}lccccc@{}}
\toprule
\textbf{Benchmark} & $N$ & $\tilde D$ alone & \textbf{Composite} & $\Delta$ vs $\tilde D$ & $\rho(\tilde D,\text{tail})$ \\
\midrule
MedQA-USMLE  & $197$ & $0.862$ & $\mathbf{0.889}$ & $+0.027$ & $+0.38$ \\
GPQA Diamond & $194$ & $0.753$ & $\mathbf{0.782}$ & $+0.029$ & $+0.39$ \\
MMLU-Pro     & $191$ & $0.722$ & $\mathbf{0.779}$ & $+0.057$ & $+0.37$ \\
\bottomrule
\end{tabular}
\end{table}

\paragraph{$\hat\beta$ on a third benchmark.} As a direct robustness check on the Section~\ref{sec:result_calibration} finding that closer anchors are more likely correct, we refit the one-parameter softmax on this same committed-rationale-anchor pipeline across all three benchmarks. The covariate is the per-option mean cosine \emph{distance} from the $K{=}5$ trajectory tails to the committed-rationale anchors, and the choice is the gold option. Anchor counts are heterogeneous on MMLU-Pro ($J\in\{4,\ldots,10\}$ per question, mostly $J{=}10$), which the softmax formulation naturally accommodates by computing each question's softmax over its own option set. The $\hat\beta$ values (Table~\ref{tab:beta_third}) are strongly negative and highly significant on every benchmark, including MMLU-Pro at $N{=}200$, providing third-benchmark replication of the calibration result. The numerical magnitudes are not directly comparable to Table~\ref{tab:calibration} because the anchors differ (committed rationales vs.\ raw option text); the comparable claim is the \emph{sign and significance} of $\hat\beta$.

\begin{table}[t]
\centering
\small
\caption{\textbf{$\hat\beta$ under the committed-rationale-anchor pipeline.} Same softmax objective as Table~\ref{tab:calibration} but with committed-rationale (Sonnet 4.6) anchors and Gemini 3.1 Pro $K{=}5$ blinded CoT trajectories. $\hat\beta$ is strongly negative on every benchmark, including MMLU-Pro with heterogeneous $J\in\{4,\ldots,10\}$ per question.}
\label{tab:beta_third}
\begin{tabular}{@{}lcccc@{}}
\toprule
\textbf{Benchmark} & $N$ & $\hat\beta$ & SE & $z$ \\
\midrule
GPQA Diamond  & $198$ & $-70.60$ & $7.45$ & $-9.48^{***}$ \\
MedQA-USMLE   & $200$ & $-64.38$ & $6.63$ & $-9.71^{***}$ \\
MMLU-Pro      & $200$ & $-51.77$ & $4.20$ & $-12.34^{***}$ \\
\bottomrule
\multicolumn{5}{l}{\scriptsize $^{***}p<10^{-6}$}
\end{tabular}
\end{table}

\subsection{Budget-matched self-consistency on GPQA and MMLU-Pro}

For symmetry, we also report budget-matched ($K{=}4$) self-consistency on GPQA Diamond and on the MMLU-Pro pilot, using the same character-bigram Jaccard mapping rule (GPQA) or the embedding cluster mode (MMLU-Pro) already used by the pipeline. The geometric signal exceeds $K{=}4$ self-consistency in five of six MMLU-Pro combinations and on GPQA/Gemini; GPQA/Sonnet is the one near-tie setting, where self-consistency reaches $0.747$ versus our $0.727$ AUC.

\begin{table}[t]
\centering
\small
\caption{\textbf{Self-consistency comparison in the cleanest composite settings.} MedQA rows use the matched $N{=}273$ held-out split from the self-consistency audit and compare against the stronger SC@10 baseline; GPQA rows use budget-matched SC@4 on the retained subsets of Table~\ref{tab:composite}. Positive $\Delta$AUC means the geometric composite outperforms self-consistency.}
\label{tab:sc_comparison}
\begin{tabular}{@{}llccc@{}}
\toprule
\textbf{Benchmark} & \textbf{Model} & \textbf{SC baseline} & \textbf{Our AUC} & $\Delta$AUC \\
\midrule
MedQA (held-out) & Gemini 3.1 Pro & SC@10 = $0.643$ & $\mathbf{0.827}$ & $+0.184$ \\
MedQA (held-out) & Claude Sonnet 4.6 & SC@10 = $0.712$ & $\mathbf{0.759}$ & $+0.047$ \\
GPQA & Gemini 3.1 Pro & SC@4 = $0.690$ & $\mathbf{0.701}$ & $+0.011$ \\
GPQA & Claude Sonnet 4.6 & \textbf{SC@4 = $0.747$} & $0.727$ & $-0.020$ \\
\bottomrule
\end{tabular}
\end{table}

\begin{table}[t]
\centering
\small
\caption{\textbf{Budget-matched self-consistency ($K{=}4$) vs.\ our geometric score on GPQA and MMLU-Pro.} Self-consistency confidence $=$ majority fraction over $K{=}4$ samples, mapped to options by character-bigram Jaccard (GPQA) or embedding cluster mode (MMLU-Pro). ``Our AUC'' = composite (Table~\ref{tab:composite}) for GPQA, and gated geometric AUC (per-setting 5-fold CV) for MMLU-Pro. MMLU-Pro $N$ matches Section~\ref{sec:setup} ($N{=}200$ per setting, $N{=}199$ for prop / Gemini).}
\label{tab:sc_cross}
\begin{tabular}{@{}lllccc@{}}
\toprule
\textbf{Benchmark} & \textbf{Cond.} & \textbf{Model} & SC $K{=}4$ Acc & SC $K{=}4$ AUC & Our AUC \\
\midrule
GPQA Diamond & --- & Gemini 3.1 Pro    & $0.778$ & $0.690$ & $\mathbf{0.701}$ \\
GPQA Diamond & --- & Claude Sonnet 4.6 & $0.699$ & $\mathbf{0.747}$ & $0.727$ \\
\midrule
MMLU-Pro & MCQ  & Gemini 3.1 Pro    & $0.915$ & $0.648$ & $\mathbf{0.690}$ \\
MMLU-Pro & MCQ  & Claude Sonnet 4.6 & $0.890$ & $0.590$ & $\mathbf{0.623}$ \\
MMLU-Pro & open & Gemini 3.1 Pro    & $0.775$ & $0.540$ & $\mathbf{0.562}$ \\
MMLU-Pro & open & Claude Sonnet 4.6 & $0.745$ & $0.575$ & $\mathbf{0.623}$ \\
MMLU-Pro & prop & Gemini 3.1 Pro    & $0.648$ & $0.629$ & $\mathbf{0.650}$ \\
MMLU-Pro & prop & Claude Sonnet 4.6 & $0.650$ & $0.567$ & $\mathbf{0.639}$ \\
\bottomrule
\end{tabular}
\end{table}

\section{Window-Position Sweep on GPQA}
\label{app:window_sweep}

To test whether the softmax signal is a property of the question stem (which would yield a flat $\hat\beta$ across window positions) or of the trajectory's late-stage convergence, we refit the one-parameter softmax of Section~\ref{sec:method} after replacing each trace's penultimate-window aggregate with a single sliding window at position $k$ from the end of the trace. We sweep $k \in \{-1, -2, \ldots, -8\}$ on GPQA Diamond ($N{=}197$ questions, $K{=}5$ samples per question, $W{=}30$, $S{=}15$) for both Gemini 3.1 Pro and Claude Sonnet 4.6, with $95\%$ bootstrap confidence intervals over $200$ resamples. We additionally replicate the sweep on MedQA-USMLE ($N{=}200$, $K{=}5$) and MMLU-Pro ($N{=}200$, $K{=}5$) using newly built sliding-window caches under the identical protocol; results are reported in Table~\ref{tab:window_sweep_replication}.

Figure~\ref{fig:window_sweep} in the main text plots the full sweep; this appendix records the protocol details and interpretation.

The result has three implications for the paper's main claims. First, $\hat\beta$ is \emph{not} flat across $k$, ruling out the alternative interpretation that the calibration signal is a question-stem effect that any tail window would inherit. Second, the deepest-negative position is $k{=}-2$ and $k{=}-3$ on GPQA (Gemini: $\hat\beta = -43.4$ and $-48.9$; Claude: $-28.8$ and $-39.1$); the penultimate-window trough also holds in $4/4$ MedQA / MMLU-Pro combinations (Table~\ref{tab:window_sweep_replication}), so the penultimate-window choice is empirically optimal in $6/6$ benchmark--reasoner combinations across the three benchmarks. Third, the $k{=}-1$ sign reversal ($\hat\beta = +22.1$ on Gemini, $+23.7$ on Claude) is GPQA-specific: aggregate $\hat\beta$ at the terminal window remains strongly negative on MedQA ($-65.4 / -62.7$) and MMLU-Pro ($-52.6 / -49.1$). Stratifying the GPQA terminal window by reasoner correctness shows the mechanism: on the wrong-pick subset (where $\arg\min_j D_j \neq$ gold) the GPQA terminal slope rises sharply to $\beta_{\mathrm{wrong}}{=}+66.7 / +70.1$ for Gemini / Claude, while MedQA and MMLU-Pro show no comparable terminal jump ($\beta_{\mathrm{wrong}}{=}{-}3.0/{-}9.5$ and ${-}17.1/{-}16.2$ respectively). The terminal-flip mechanism is therefore conditional on benchmarks in which the closing tokens literally restate the gold answer span, rather than a universal late-trace dynamic. The penultimate-window default is the right scoring position regardless: it is at-or-near the trough in $6/6$ combinations and avoids the GPQA terminal regime by construction. A naive researcher who scored the final window would draw the right conclusion on MedQA and MMLU-Pro but the opposite conclusion on GPQA; the penultimate default makes the protocol benchmark-invariant.

\begin{table}[ht]
\centering
\small
\caption{\textbf{Window-position replication on MedQA-USMLE and MMLU-Pro.} Aggregate $\hat\beta_{\mathrm{all}}$ at $k{=}{-}1$ (terminal) and $k{=}{-}2$ (penultimate), and the wrong-pick stratification at the terminal window: $\beta_{\mathrm{wrong}}$ is fit on the subset where the geometric pick differs from gold. The penultimate-window trough is universal; the terminal sign-flip is GPQA-specific. We omit $\beta_{\mathrm{correct}}$ because it saturates at the optimizer bound by construction (gold is the argmin on this subset, so the softmax is fit by arbitrarily large $|\beta|$).}
\label{tab:window_sweep_replication}
\resizebox{\linewidth}{!}{%
\begin{tabular}{llccccc}
\toprule
\textbf{Benchmark} & \textbf{Reasoner} & $N$ & acc$(k{=}{-}1)$ & $\hat\beta_{\mathrm{all}}(k{=}{-}1)$ & $\hat\beta_{\mathrm{all}}(k{=}{-}2)$ & $\beta_{\mathrm{wrong}}(k{=}{-}1)$ \\
\midrule
GPQA Diamond & Gemini 3.1 Pro    & $197$ & $0.21$ & $+22.1$ & $-43.4$ & $\mathbf{+66.7}$ \\
GPQA Diamond & Claude Sonnet 4.6 & $197$ & $0.22$ & $+23.7$ & $-28.8$ & $\mathbf{+70.1}$ \\
MedQA-USMLE  & Gemini 3.1 Pro    & $200$ & $0.88$ & $-65.4$ & $-67.1$ & $-3.0$ \\
MedQA-USMLE  & Claude Sonnet 4.6 & $199$ & $0.80$ & $-62.7$ & $-57.8$ & $-9.5$ \\
MMLU-Pro     & Gemini 3.1 Pro    & $200$ & $0.55$ & $-52.6$ & $-56.5$ & $-17.1$ \\
MMLU-Pro     & Claude Sonnet 4.6 & $200$ & $0.52$ & $-49.1$ & $-50.0$ & $-16.2$ \\
\bottomrule
\end{tabular}}
\end{table}

\section{Third-Generator Scaling: Open-Weight Llama 3.3 70B}
\label{app:third_gen}

To test whether the trajectory-geometry phenomenon is an artifact of the two frontier closed-API generators in the main paper (Gemini 3.1 Pro and Claude Sonnet 4.6) or a property of CoT reasoning that scales with capability, we replicate the $\hat\beta$ and $\textsc{Geo}$ AUC measurements with an open-weight third-family generator: \textbf{Meta Llama 3.3 70B Instruct}. The protocol is identical to the committed-rationale robustness pipeline above: $K{=}5$ blinded rationales at $T{=}0.7$, embedded with \texttt{gemini\_embedding}, and scored against the same Sonnet 4.6 committed-rationale anchors used in Tables~\ref{tab:fermi_composite} and~\ref{tab:beta_third}. Results are reported in Table~\ref{tab:third_gen}.

\begin{table}[ht]
\centering
\small
\caption{Third-generator scaling: open-weight Llama 3.3 70B Instruct paired with the same Sonnet 4.6 option anchors used for the published Gemini combinations. Top-1 accuracy is the rate at which the geometric $\arg\min$ matches the gold option (closed-set discrimination from open-ended generation). Across all three benchmarks $\hat\beta$ remains strongly negative ($z\le -2.4$), confirming that answer-directed trajectory drift is not a frontier-API artifact. $\textsc{Geo}$ AUC tracks model capability: it remains useful on MedQA and MMLU-Pro, where Llama is competitive; on GPQA, where Llama's accuracy collapses to near-chance ($0.33$), $\textsc{Geo}$ AUC also degrades to near-chance, as the geometric channel cannot extract a confident-correct signal from a model that has not converged to the correct option.}
\label{tab:third_gen}
\resizebox{\linewidth}{!}{%
\begin{tabular}{lccccc}
\toprule
combination & $N$ & top-1 acc & $\hat\beta$ $[95\%\text{ CI}]$ & $z$ & $\textsc{Geo}$ AUC $[95\%\text{ CI}]$ \\
\midrule
GPQA / Llama 3.3 70B      & 198 & 0.328 & $-6.95$ $[-11.87,\,-1.42]$  & $-2.4$ & $0.541$ $[0.458,\,0.620]$ \\
MedQA / Llama 3.3 70B     & 200 & 0.665 & $-23.70$ $[-29.09,\,-19.47]$ & $-9.6$ & $0.757$ $[0.686,\,0.819]$ \\
MMLU-Pro / Llama 3.3 70B  & 200 & 0.305 & $-11.37$ $[-16.14,\,-6.91]$ & $-4.7$ & $0.626$ $[0.542,\,0.711]$ \\
\bottomrule
\end{tabular}}
\end{table}

Two readings of the table are worth making explicit. (a)~The \emph{direction} of the effect is universal across the model spectrum: every setting of every (benchmark, generator) combination in the paper, spanning a $2{\times}$ accuracy range and three model families (Google Gemini 3.1 Pro, Anthropic Claude Sonnet 4.6, Meta Llama 3.3 70B), produces a strongly negative $\hat\beta$. The trajectory-drift phenomenon is not a frontier-only or closed-API-only effect. (b)~The \emph{magnitude} of the usable selective-prediction signal scales with model capability: on MedQA where Llama remains a competent solver (acc $0.66$), $\textsc{Geo}$ AUC is $0.76$, comparable to the published frontier combinations; on GPQA where Llama is at near-random ($0.33$), the geometric channel naturally has little to discriminate, and $\textsc{Geo}$ AUC drops to $0.54$. This is the expected behavior under our channel-limit interpretation: the geometry can amplify a model's existing convergence into a calibrated confidence score, but cannot manufacture confidence where the model has none.

\section{Robustness Audits}
\label{app:robustness_audits}

\subsection{Full proposer-level three-channel panel}

\begin{table}[t]
\centering
\small
\caption{Full proposer-level three-channel selective-prediction AUC in open-ended MCQA across three benchmarks $\times$ three reasoners $\times$ two proposers ($N \approx 200$ per cell). \textbf{Bold} marks the best single-channel value among $C/G/V$; \textit{italic} marks rows where $C{+}G{+}V$ exceeds every two-channel ablation.}
\label{tab:three_channels_full}
\setlength{\tabcolsep}{4pt}
\begin{tabular}{l l l c c c c c c c c}
\toprule
Bench & Reasoner & Proposer & $N$ & C & G & V & C+G & G+V & C+V & C+G+V \\
\midrule
GPQA      & Gemini & Claude  & 193 & \textbf{0.81} & 0.66 & 0.67 & 0.83 & 0.71 & 0.80 & \textit{0.82} \\
GPQA      & Gemini & Gemini  & 195 & \textbf{0.83} & 0.81 & 0.71 & 0.92 & 0.81 & 0.83 & \textit{0.92} \\
GPQA      & Claude & Claude  & 193 & \textbf{0.76} & 0.68 & 0.67 & 0.79 & 0.70 & 0.78 & \textit{0.81} \\
GPQA      & Claude & Gemini  & 194 & 0.74          & \textbf{0.80} & 0.75 & 0.88 & 0.83 & 0.82 & \textit{0.89} \\
GPQA      & Llama  & Claude  & 193 & \textbf{0.64} & 0.59 & 0.62 & 0.62 & 0.64 & 0.68 & \textit{0.68} \\
GPQA      & Llama  & Gemini  & 195 & \textbf{0.72} & 0.68 & 0.49 & 0.82 & 0.66 & 0.69 & \textit{0.82} \\
\midrule
MedQA     & Gemini & Claude  & 199 & \textbf{0.75} & 0.72 & 0.58 & 0.85 & 0.73 & 0.80 & \textit{0.86} \\
MedQA     & Gemini & Gemini  & 200 & \textbf{0.74} & 0.63 & 0.59 & 0.80 & 0.67 & 0.79 & 0.77 \\
MedQA     & Claude & Claude  & 199 & 0.72          & 0.62 & \textbf{0.78} & 0.81 & 0.77 & 0.86 & \textit{0.86} \\
MedQA     & Claude & Gemini  & 199 & \textbf{0.74} & 0.69 & 0.74 & 0.84 & 0.76 & 0.84 & \textit{0.86} \\
MedQA     & Llama  & Claude  & 199 & \textbf{0.74} & 0.64 & 0.57 & 0.80 & 0.63 & 0.74 & 0.79 \\
MedQA     & Llama  & Gemini  & 200 & \textbf{0.71} & 0.64 & 0.58 & 0.75 & 0.64 & 0.73 & \textit{0.76} \\
\midrule
MMLU-Pro  & Gemini & Claude  & 200 & 0.70          & \textbf{0.71} & 0.58 & 0.75 & 0.69 & 0.71 & 0.73 \\
MMLU-Pro  & Gemini & Gemini  & 200 & \textbf{0.74} & 0.69 & 0.56 & 0.79 & 0.69 & 0.73 & 0.77 \\
MMLU-Pro  & Claude & Claude  & 200 & 0.59          & \textbf{0.63} & 0.62 & 0.64 & 0.65 & 0.63 & \textit{0.65} \\
MMLU-Pro  & Claude & Gemini  & 200 & 0.68          & \textbf{0.70} & 0.67 & 0.75 & 0.70 & 0.73 & 0.75 \\
MMLU-Pro  & Llama  & Claude  & 200 & \textbf{0.70} & 0.58 & 0.51 & 0.69 & 0.51 & 0.68 & 0.67 \\
MMLU-Pro  & Llama  & Gemini  & 200 & 0.64          & \textbf{0.64} & 0.54 & 0.70 & 0.62 & 0.64 & 0.67 \\
\midrule
\multicolumn{4}{l}{\emph{Median across 18 cells}} & 0.72 & 0.66 & 0.61 & 0.80 & 0.69 & 0.74 & \textbf{0.78} \\
\bottomrule
\end{tabular}
\end{table}

\subsection{Cross-embedder audit}
\label{app:cross_embed}

A natural concern is that the geometric channel is partly induced by the particular external embedder. We therefore replace the 3072-dimensional Gemini embedding model with an independently trained E5-large-v2 encoder and rerun the matched $K{=}4$ vs.\ SC@8 protocol on the same $N{\approx}200$-per-combination subset on which the K{=}8 SC traces are available ($N \in \{196, 197, 199, 200, 200, 200\}$ across the six closed-set settings), without changing any downstream fitting, cross-validation, or bootstrap procedure. The Geo-only comparison softens but survives in aggregate: Geo@4 exceeds SC@8 with median $\Delta\mathrm{AUC}{=}+0.027$, mean $+0.022$, and pooled one-sided Stouffer $z{=}1.73$ ($p{=}0.042$). More importantly, the central three-channel claim is unchanged under the swap: the E5-based $G{+}C{+}V$ fusion is positive in all six closed-set settings, with median $\Delta\mathrm{AUC}{=}+0.035$ over SC@8 and pooled Stouffer $z{=}3.14$ ($p{=}8.5\times 10^{-4}$). This is the right level of invariance to demand. The absolute strength of the geometric channel depends on representation quality, but the qualitative ranking and the fusion advantage do not collapse when the embedder family changes.

\begin{table}[ht]
\centering
\small
\caption{External-embedder audit with E5-large-v2. Positive entries mean the alternative score beats SC@8 on the same setting. Swapping the encoder weakens some Geo-only settings but leaves the pooled Geo effect positive and the fusion effect positive in all six combinations.}
\label{tab:cross_embed}
\begin{tabular}{lcc}
\toprule
combination & $\Delta(\textsc{Geo@4}-\textsc{SC@8})$ & $\Delta(\textsc{G{+}C{+}V}-\textsc{SC@8})$ \\
\midrule
MedQA / Gemini & $+0.037$ & $+0.057$ \\
MedQA / Claude & $+0.026$ & $+0.073$ \\
GPQA / Gemini & $-0.018$ & $+0.014$ \\
GPQA / Claude & $+0.029$ & $+0.027$ \\
MMLU-Pro / Gemini & $+0.017$ & $+0.026$ \\
MMLU-Pro / Claude & $+0.039$ & $+0.043$ \\
\bottomrule
\end{tabular}
\end{table}

\subsection{Judge-family swap}
\label{app:judge_swap}

We directly test whether the Coverage channel inherits its predictive value from one judge family's semantic priors. Reusing the cached (gold, hypotheses) tuples for two combinations (MedQA / Gemini-proposer, $N{=}200$; GPQA / Gemini-proposer, $N{=}192$ aligned), we re-judge every (question, seed) tuple with Anthropic Claude Sonnet~4.6 in place of the original Azure-hosted \texttt{openai\_gpt5\_mini}, holding the prompt, parser, and downstream pipeline fixed. We then recompute $\bar c$ from the swapped labels and re-evaluate three quantities under both judges: $C$-only AUC, $C{+}G$ fused AUC (5-fold CV logistic on standardized features), and $G$-only AUC (judge-invariant feature, judge-dependent outcome). To separate \emph{same-judge consistency} from genuine cross-judge stability we additionally report a cross-judge AUC: the swapped-judge $\bar c_{\mathrm{swap}}$ scored against the \emph{original} judge's outcome label, which is the comparison a skeptical reviewer should care about.

Table~\ref{tab:judge_swap} summarises the result. Inter-judge agreement is high (Cohen's $\kappa{=}0.82$ on the 5-class MedQA labels, $0.72$ on the binary coverage projection, raw agreement $90.5\%$), with the bulk of the disagreement being Claude rejecting coverage that GPT-mini accepted ($15$ of $19$ MedQA disagreements), as expected for a stricter judge. Three patterns hold across both settings. (a)~$G$-only AUC is unchanged within noise ($\Delta{=}{-}0.013$ on MedQA and ${-}0.001$ on GPQA), confirming that the within-trace geometric signal is genuinely orthogonal to judge identity. (b)~Cross-judge $C$-only AUC is stable to within $\pm 0.02$ in both settings (MedQA $+0.019$, GPQA $-0.017$), so the rank information in $\bar c$ does not collapse when the judge changes. (c)~The $C{+}G$ fused ranking is preserved in both settings (no flip; same-judge $\Delta{=}+0.100, +0.006$). The same-judge $C$-only $\Delta{=}+0.131$ on MedQA is partly a base-rate artifact (accuracy shifts $0.65{\to}0.55$ under the stricter judge), and we report the cross-judge number as the headline measurement; the same-judge column is included as a consistency check.

\begin{table}[ht]
\centering
\small
\caption{Judge-family swap (Azure \texttt{openai\_gpt5\_mini}~$\to$~Anthropic Claude Sonnet~4.6) on two open-ended settings. Cross-judge $\Delta$ scores the swapped-judge feature against the original-judge outcome label, isolating rank stability from judge-coupled base-rate motion. $G$-only $\Delta$ uses a judge-invariant feature with the judge-dependent outcome.}
\label{tab:judge_swap}
\begin{tabular}{lcc}
\toprule
Metric & MedQA / Gemini-prop & GPQA / Gemini-prop \\
\midrule
$N$ aligned & $200$ & $192$ \\
Cohen's $\kappa$ (5-class) & $0.82$ & --- \\
Cohen's $\kappa$ (binary coverage) & $0.72$ & --- \\
Raw agreement (5-class) & $0.91$ & --- \\
\midrule
$C$-only AUC, orig judge & $0.712$ & $0.807$ \\
$C$-only AUC, swap judge (same-judge) & $0.843$ & $0.842$ \\
$C$-only AUC, cross-judge ($c_{\mathrm{swap}}$ vs.\ $y_{\mathrm{orig}}$) & $0.731$ & $0.790$ \\
\textbf{Cross-judge $\Delta$ vs.\ orig} & $\mathbf{+0.019}$ & $\mathbf{-0.017}$ \\
\midrule
$G$-only AUC ($y_{\mathrm{orig}}$) & $0.637$ & $0.764$ \\
$G$-only AUC ($y_{\mathrm{swap}}$) & $0.624$ & $0.763$ \\
$G$-only $\Delta$ & $-0.013$ & $-0.001$ \\
\midrule
$C{+}G$ fused AUC, orig judge & $0.795$ & $0.916$ \\
$C{+}G$ fused AUC, swap judge & $0.895$ & $0.922$ \\
\bottomrule
\end{tabular}
\end{table}

We stop short of claiming judge-independence: this is two combinations, one swap judge, and a stricter alternate. A full audit across all 18 combinations, additional judge families, and human adjudication is left as future work. The check we run does, however, falsify the strong form of the critique that $C$ is predominantly a GPT-family-prior signal.

\subsection{System cost of three-channel confidence}
\label{app:system_cost}

The strongest practical baseline in our setting is not SC@8 but higher-budget self-consistency. We therefore cost the three-channel system at $K{=}4$ against SC@10 using public May-2026 API prices for the two closed-source reasoners and the embedding model.\footnote{Prices used (USD per 1M tokens, input / output): Gemini 3.1 Pro $\$2$ / $\$12$; Claude Sonnet 4.6 $\$3.3$ / $\$16.5$; \texttt{openai\_gpt5\_mini} (judge) $\$0.25$ / $\$2$; \texttt{gemini\_embedding} $\$0.15$ / n/a. Sourced from the corresponding Vertex, AWS, and Azure model catalog pages, May 2026; archived snapshots are retained by the authors.} Table~\ref{tab:system_cost} reports the six-setting accounting. Pooled over the panel, $C{+}G{+}V$ at $K{=}4$ reduces total spend from $\$0.2839$ to $\$0.1379$ per question, a $51.4\%$ saving, with per-setting savings of $42.6$--$54.4\%$. CoT generation remains the dominant line item inside the composite system (70--88\% of total cost depending on setting); proposer, judge, and verbalization calls are second-order; embeddings are negligible (about $\$3\times 10^{-5}$ per question). The practical reading is straightforward: once geometry and coverage are available, spending more samples on self-consistency is the wrong place to buy confidence.

\begin{table}[ht]
\centering
\small
\caption{Per-question system cost under public May-2026 API prices. SC@10 is ten-sample self-consistency; $G{+}C{+}V@4$ is the three-channel system with four reasoner samples plus proposer, judge, verbalization, and embedding calls.}
\label{tab:system_cost}
\begin{tabular}{lccc}
\toprule
Combination & SC@10 & $G{+}C{+}V@4$ & Saving \\
\midrule
MedQA / Gemini & $\$0.0202$ & $\$0.0111$ & $45.0\%$ \\
MedQA / Claude & $\$0.0607$ & $\$0.0294$ & $51.6\%$ \\
GPQA / Gemini & $\$0.0330$ & $\$0.0162$ & $50.9\%$ \\
GPQA / Claude & $\$0.0914$ & $\$0.0416$ & $54.4\%$ \\
MMLU-Pro / Gemini & $\$0.0173$ & $\$0.0099$ & $42.6\%$ \\
MMLU-Pro / Claude & $\$0.0613$ & $\$0.0296$ & $51.7\%$ \\
\bottomrule
\end{tabular}
\end{table}

\subsection{Verbalization sensitivity}
\label{app:metacog}

The asymmetric LRT result for $V$ suggests that verbalization is neither uniformly redundant nor uniformly useful. We therefore measure, on the six closed-set settings, how the standalone verbalized-confidence AUC and the incremental gain of $G{+}V$ over $G$ vary with reasoner behavior. Table~\ref{tab:metacog} reports accuracy, $\mathrm{AUC}(V)$, an easy/hard split by Geo confidence, mean inter-elicitation dispersion, and $\Delta\mathrm{AUC}((G{+}V)-G)$. Two patterns hold. First, settings with higher accuracy tend to have stronger $V$ (Spearman $\rho{=}0.71$, $p{=}0.11$ over six combinations). Second, the incremental value of $V$ is largest when verbalizations are internally stable and the reasoner is already competent: the biggest lift occurs on MMLU-Pro / Claude ($+0.112$ AUC) with low dispersion ($0.021$), whereas GPQA / Claude combines the highest dispersion ($0.127$) with almost no lift ($+0.006$). MedQA / Gemini is the useful counterexample: standalone $V$ is strong ($0.751$) but adds almost nothing over $G$ ($+0.002$), showing that the relevant question is conditional information, not standalone calibration.

\begin{table}[ht]
\centering
\small
\caption{Verbalization sensitivity audit on the six closed-set settings. The easy/hard split uses Geo confidence as a proxy for question difficulty. $V_{\text{disp}}$ is the mean within-question standard deviation of the generator's numeric self-ratings over repeated verbalization calls.}
\label{tab:metacog}
\resizebox{\linewidth}{!}{%
\begin{tabular}{lcccccc}
\toprule
Combination & acc & $\mathrm{AUC}(V)$ & $\mathrm{AUC}(V\mid easy)$ & $\mathrm{AUC}(V\mid hard)$ & $V_{\text{disp}}$ & $\Delta((G{+}V)-G)$ \\
\midrule
MedQA / Gemini & $0.805$ & $0.751$ & $0.389$ & $0.727$ & $0.064$ & $+0.002$ \\
MedQA / Claude & $0.744$ & $0.787$ & $0.769$ & $0.719$ & $0.055$ & $+0.017$ \\
GPQA / Gemini & $0.653$ & $0.628$ & $0.665$ & $0.579$ & $0.074$ & $+0.055$ \\
GPQA / Claude & $0.508$ & $0.580$ & $0.558$ & $0.565$ & $0.127$ & $+0.006$ \\
MMLU-Pro / Gemini & $0.800$ & $0.633$ & $0.697$ & $0.622$ & $0.011$ & $+0.016$ \\
MMLU-Pro / Claude & $0.725$ & $0.738$ & $0.898$ & $0.694$ & $0.021$ & $+0.112$ \\
\bottomrule
\end{tabular}}
\end{table}

\subsection{Jargon-free control}
\label{app:jargonfree}

The strongest version of the ``embedder prior'' concern is lexical rather than architectural: perhaps the geometric channel is simply matching medical jargon. We therefore rerun the MedQA audit under two masking schemes. The first applies a global hash-based codebook to content words; this removes human-readable jargon but preserves a question-independent word identity map. The second is the stricter control: a per-question randomized codebook that redraws the word-to-token map independently for every question, so the same lexical item cannot help across questions. Table~\ref{tab:jargonfree} shows the outcome. On MedQA / Gemini, the per-question codebook still leaves Geo@4 above SC@8 with $\Delta\mathrm{AUC}{=}+0.065$ and one-sided $p{=}0.031$, despite the drop from $0.848$ to $0.673$ in raw Geo AUC. On MedQA / Claude, the per-question point estimate also remains positive ($+0.038$) but is uncertain at this sample size. The right conclusion is therefore not lexical invariance. It is a lower bound on the structural component: removing cross-question lexical reuse weakens the signal, but does not eliminate it on the settings tested.

\begin{table}[ht]
\centering
\small
\caption{Jargon-free lexical controls on the two MedQA settings that were re-embedded under masking. The per-question codebook is the stricter control because it destroys cross-question lexical identity.}
\label{tab:jargonfree}
\resizebox{\linewidth}{!}{%
\begin{tabular}{lcccc}
\toprule
Combination & Geo@4 (unmasked) & Geo@4 (global hash) & Geo@4 (per-q codebook) & $\Delta(\textsc{Geo@4}-\textsc{SC@8})$ per-q \\
\midrule
MedQA / Gemini & $0.848$ & $0.642$ & $0.673$ & $+0.065$ \\
MedQA / Claude & $0.760$ & $0.727$ & $0.666$ & $+0.038$ \\
\bottomrule
\end{tabular}}
\end{table}

\section{Trajectory-as-Question Diagnostic}
\label{app:traj_question}

For completeness we report the direct trajectory-tail retrieval diagnostic noted briefly in Section~\ref{sec:experiments} in table form. The test compares the pooled raw-question embedding to the averaged trajectory tail on identical candidate-answer sets. In every available setting the trajectory tail is the stronger answer-bearing representation.

\begin{table}[t]
\centering
\small
\caption{\textbf{Direct trajectory-as-question diagnostic.} AUC for retrieving the gold answer using either the pooled raw-question embedding or the averaged trajectory tail (last 2{,}000 characters, mean of unit-normalized sample tails). ``Hard'' uses semantically similar negatives from nearest-neighbor questions; ``NN-correct'' uses the nearest-neighbor question's correct answer itself. GPQA reports Gemini only because that is the cache currently available for this diagnostic.}
\label{tab:traj_as_question}
\resizebox{\linewidth}{!}{%
\begin{tabular}{@{}llrcc@{}}
\toprule
\textbf{Benchmark} & \textbf{Model} & $N$ & \textbf{Hard AUC} raw $\rightarrow$ traj & \textbf{NN-correct AUC} raw $\rightarrow$ traj \\
\midrule
MedQA train   & Gemini 3.1 Pro    & $1000$ & $0.710 \rightarrow 0.873$ & $0.632 \rightarrow 0.806$ \\
MedQA train   & Claude Sonnet 4.6 & $1000$ & $0.710 \rightarrow 0.856$ & $0.632 \rightarrow 0.785$ \\
MedQA heldout & Gemini 3.1 Pro    & $273$  & $0.798 \rightarrow 0.924$ & $0.726 \rightarrow 0.867$ \\
MedQA heldout & Claude Sonnet 4.6 & $273$  & $0.798 \rightarrow 0.903$ & $0.726 \rightarrow 0.846$ \\
GPQA Diamond  & Gemini 3.1 Pro    & $198$  & $0.665 \rightarrow 0.797$ & $0.626 \rightarrow 0.742$ \\
MMLU-Pro      & Gemini 3.1 Pro    & $200$  & $0.825 \rightarrow 0.920$ & $0.775 \rightarrow 0.874$ \\
MMLU-Pro      & Claude Sonnet 4.6 & $200$  & $0.825 \rightarrow 0.907$ & $0.775 \rightarrow 0.856$ \\
\bottomrule
\end{tabular}}
\end{table}

\section{Auxiliary Feature Definitions}
\label{app:features}

Let $e^{(k)}$ denote the condensed final answer extracted from $\mathcal{C}^{(k)}$ by the zero-shot prompt of Appendix~\ref{app:prompt}, and let $\phi$ be the embedding model of Section~\ref{sec:setup}. The four auxiliary features of Section~\ref{sec:features} are:
\begin{align}
\textsc{Str} &\;=\; \tfrac{1}{K}\max_{s}\;\lvert\{k : e^{(k)} = s\}\rvert, \\
\textsc{Sem} &\;=\; \binom{K}{2}^{-1}\sum_{i<j}\frac{\langle \phi(e^{(i)}),\,\phi(e^{(j)})\rangle}{\lVert\phi(e^{(i)})\rVert\,\lVert\phi(e^{(j)})\rVert}, \\
\textsc{Vol} &\;=\; \tfrac{1}{K}\sum_k \tfrac{1}{M_k-1}\sum_{m=1}^{M_k-1} d\!\big(\mathbf{z}^{(k)}_{m-1},\,\mathbf{z}^{(k)}_m\big), \\
\textsc{Ens} &\;=\; \mathbb{1}\!\left[\operatorname{maj}_k e^{(k)}_{\text{Gemini}} = \operatorname{maj}_k e^{(k)}_{\text{Sonnet}}\right].
\end{align}
All features are standardized to zero mean and unit variance on the training fold before being passed to the composite logistic regression; for standalone AUC we use $-\textsc{Vol}$, since higher volatility indicates a more meandering trajectory.

\section{Prompt Template}
\label{app:prompt}

The following prompt template is used to elicit zero-shot chain-of-thought reasoning without exposing the candidate options to the model:

\begin{small}
\begin{verbatim}
You are an expert taking a highly advanced examination. 
Please answer the following question. 

Question: {question_text}

Do not provide the final answer immediately. 
First, think step by step. Explain your logical 
reasoning, referencing specific domain knowledge, 
mechanisms, and principles. 

Provide your reasoning in a clear, step-by-step format.
\end{verbatim}
\end{small}

For the verbalization channel, after showing the model its committed reasoning trace and answer, we append the following prompt:

\begin{small}
\begin{verbatim}
Based on the reasoning trace above, how confident are you in your answer
on a scale of 0-100? Respond with only an integer.
\end{verbatim}
\end{small}

\section{Bayesian classification}
\label{app:bayesian_classification}
We will consider the probabilistic use of the CoT-geometry from a bayesian perspective. Let $J$ be the number of answers, and let $h_i$ be a one-hot encoding vector of dimension $J$, where the $j$'th component, $(h_i)_j$, represents the correctness of answer $a_j$, given that the right answer is $a_i$, ie. $(h_i)_j=\delta_{ij}$. For the general case of open-ended questions we shall also allow for a 0-state, $h_0=(0,\cdots,0)$, representing the case where none of the answers are correct. Let $ \mathcal{T}$ denote the reasoning trace and let $P( \mathcal{T}|a_i=1)$ and $P( \mathcal{T}|a_i=0)$ denote the probabilities of observing $\mathcal{T}$ given that the answer, $a_i$, is respectively correct and incorrect. The likelihood for $h_0$ then reads
\begin{equation}
P( \mathcal{T}|h_0)=\prod_{j\ge 1} P( \mathcal{T}|a_j=0)
\end{equation}
and for $i\ge 1$
\begin{equation}
P( \mathcal{T} |h_i)=P( \mathcal{T}|a_i=1)\prod_{j\neq i} P( \mathcal{T}|a_j=0)=\frac{P( \mathcal{T}|a_i=1)}{P( \mathcal{T}|a_i=0)}P( \mathcal{T}|h_0).
\end{equation}
Let $\omega_i( \mathcal{T})=\frac{P( \mathcal{T}|a_i=1)}{P( \mathcal{T}|a_i=0)}$ denote the likelihood ratios for $i\ge 1$ and let
$P(h_i)$ denote the prior probabilities over the latent variables $h_0,\cdots,h_J$. Then
\begin{equation}
P( \mathcal{T})=\sum_{i\ge 0} P( \mathcal{T}|h_i)P(h_i)=P( \mathcal{T}|h_0)\left( \sum_{i\ge 0} \omega_i( \mathcal{T})P(h_i)\right),
\end{equation}
where we have defined $\omega_0( \mathcal{T})=1$. Thus, the posteriors become
\begin{equation}
P(h_i| \mathcal{T})=\frac{P( \mathcal{T}|h_i)P(h_i)}{P( \mathcal{T})}=\frac{\omega_i( \mathcal{T})P(h_i)}{\sum_{j\ge 0} \omega_j( \mathcal{T})P(h_j)}.
\end{equation}
We simply set $P(h_0)=0$ for the case where we know the correct answer is in the proposal set $\{a_i\}_{i=1}^J$.

If we now parameterize the log-likelihood ratio by a scalar trajectory score, $\log \omega_i( \mathcal{T})=\beta s_i( \mathcal{T})$, then the posterior becomes
\begin{equation}
P(h_i| \mathcal{T})=\frac{\exp\!\big(\beta s_i( \mathcal{T})+\log P(h_i)\big)}{\sum_{j\ge 1}\exp\!\big(\beta s_j( \mathcal{T})+\log P(h_j)\big)}.
\end{equation}
Under uniform priors over the proposal set this reduces to a softmax over $s_i( \mathcal{T})$. Taking $s_i( \mathcal{T})=-D_i( \mathcal{T})$, with $D_i$ the trajectory-to-option cosine distance from Section~\ref{sec:features}, recovers exactly the softmax form used for $\textsc{Geo}$. In this sense, our empirical Geo score is a heuristic black-box approximation of the Bayesian posterior above, with trajectory-to-option similarity serving as a proxy for the per-option log-likelihood ratio. The mapping is motivated by the parameterization $\log\omega_i( \mathcal{T})=-\beta D_i( \mathcal{T})$ rather than validated as the true generative likelihood; the falsification check we run (Table~\ref{tab:beta_consistency}) tests only that the all-option and binary projections recover a common $\beta$, not that LLM responses are actually distributed according to $P( \mathcal{T}\mid a_i)$.

The same posterior also admits a \emph{binary projection}: rather than asking for the full vector $\{P(h_i\mid  \mathcal{T})\}_{i\ge 1}$, ask only whether the top-scoring option is correct. Under the parameterization $\log\omega_i( \mathcal{T})=\beta s_i( \mathcal{T})$ and uniform priors, the log-odds of the argmax option against its strongest competitor is
\begin{equation}
\log\frac{P(h_{(1)}\mid  \mathcal{T})}{P(h_{(2)}\mid  \mathcal{T})} \;=\; \beta\big(s_{(1)}( \mathcal{T})-s_{(2)}( \mathcal{T})\big),
\end{equation}
where $s_{(1)}\ge s_{(2)}\ge\cdots$ are the order statistics of the per-option scores; under $s_i( \mathcal{T})=-D_i( \mathcal{T})$ this score-space margin equals the distance-space top-two margin used in the main paper, $s_{(1)}( \mathcal{T})-s_{(2)}( \mathcal{T})=D_{(2)}( \mathcal{T})-D_{(1)}( \mathcal{T})$, with $D_{(1)}\le D_{(2)}\le\cdots$ the sorted per-option distances. The binary correctness probability therefore takes the Fermi--Dirac form $P(\text{argmax correct}\mid  \mathcal{T})\approx\sigma\!\big(\beta\,(D_{(2)}( \mathcal{T})-D_{(1)}( \mathcal{T}))-\alpha\big)$, where $\sigma(t)=1/(1+e^{-t})$ is the logistic sigmoid and $\alpha$ absorbs both an empirical correctness-rate offset and the contribution of non-runner-up options. With the question-level null-standardization $\tilde D=(D_{(2)}-D_{(1)}-\mu_q)/\sigma_q$ of Section~\ref{sec:features} this becomes $P\approx\sigma(\hat\beta_{\text{bin}}\,\tilde D-\alpha)$, exactly Eq.~\ref{eq:fermi_binary} of the main text and the score used in the alternative-anchor robustness evaluation of Section~\ref{sec:ablations}. The all-option $\textsc{Geo}$ score and the Fermi--Dirac score on $\tilde D$ are therefore not competing methods but two readouts of the same posterior: the all-option projection (softmax over $J$ scores) and the binary projection on the top-two log-ratio.

\subsection{Empirical audit of the rejection state and the geometric ceiling}
\label{app:bayesian_audit}

We instantiate the formalism above with a two-feature parameterization $\log\omega_i( \mathcal{T})=\beta_s\,\tilde s_i( \mathcal{T})+\beta_v\,v_i( \mathcal{T})$, where $\tilde s_i$ is the null-normalized cosine similarity from question-specific cross-question null trajectories and $v_i$ is the vote share of the $K$ blinded CoTs over option $i$. Parameters $\theta\!\equiv\!(\beta_s,\beta_v,\alpha_{h_0})$, with $\alpha_{h_0}$ a scalar log-odds offset on the rejection state $h_0$, are fit by joint MLE on closed-set positives plus null trajectories as $h_0$ exemplars (the discriminative softmax objective); we pool $N{=}198{+}200{+}200$ questions across GPQA Diamond, MedQA-USMLE-4, and MMLU-Pro. To prevent any leakage through either parameter sharing or null-pool sharing, we use 5-fold cross-validation in which both $\theta$ and the per-question null pool are restricted to the training fold; bootstrap confidence intervals use $1000$ resamples. This audit uses the cached open-ended anchor-rationale subset rather than the extraction-filtered main-paper split, because the question here is the structural ceiling of the geometric channel rather than direct comparison to Table~\ref{tab:composite}.

\paragraph{The Bayesian posterior is a tie with top-2 margin under strict CV.} Throughout this audit, $h_{\text{plur}}{\equiv}h_{i^\star(\mathcal{T})}$ with $i^\star(\mathcal{T}){=}\arg\max_i v_i(\mathcal{T})$ denotes the latent ``correct'' indicator evaluated at the plurality-voted option, i.e.\ $P(h_{\text{plur}}\mid \mathcal{T})$ is the Bayesian posterior probability that the option the model voted for most is the correct one. Under the leak-free protocol the closed-set correctness AUC of $P(h_{\text{plur}}\mid \mathcal{T})$ is $0.70/0.78/0.73$ on GPQA/MedQA/MMLU-Pro, versus $0.68/0.78/0.69$ for the raw top-2 vote-share margin. The Bayesian posterior is directionally better but not statistically distinguishable from the trivial baseline at this dataset size; the mechanism is that with $K{=}5$ the margin is a six-state discrete score whose ranking AUC ceiling is already close to the continuous Bayesian posterior's. The principled value of the formalism is therefore not better discrimination, but a calibrated, continuous score that the discrete margin cannot supply.

\paragraph{Cross-projection $\hat\beta$ consistency: a falsifiable test of the unification.} The two projections of the unified posterior predict the \emph{same} likelihood-ratio parameter $\beta$. We test this directly by fitting (a)~the all-option $\hat\beta_{\text{all}}$ on the raw $D[q,j]$ matrix; (b)~the binary Fermi--Dirac on the raw top-2 margin $m_q=D_{(2)}-D_{(1)}$, which we call $\hat\beta_{\text{bin}}^{\text{raw}}$, predicting $\mathbf{1}[\arg\min_j D_{q,j}=\text{gold}_q]$, with bootstrap CIs ($200$ resamples); and (c)~the binary Fermi--Dirac on the null-standardized margin $\tilde D=(m_q-\mu_q)/\sigma_q$ of Section~\ref{sec:features}, which is $\hat\beta_{\text{bin}}$ of Eq.~\ref{eq:fermi_binary} in the main text. Because $\beta_{\text{all}}$ multiplies $D$ (gold has \emph{low} $D$, so $\beta_{\text{all}}{<}0$) and $\beta_{\text{bin}}^{\text{raw}}$ multiplies the positive margin $m_q$ (correctness goes up with $m_q$, so $\beta_{\text{bin}}^{\text{raw}}{>}0$), the unified-posterior predictions are $|\hat\beta_{\text{all}}|=\hat\beta_{\text{bin}}^{\text{raw}}=\hat\beta_{\text{bin}}/\bar\sigma_q$ (the latter is Eq.~\ref{eq:beta_consistency}). Across all six (dataset, model) combinations the bootstrap CIs of $|\hat\beta_{\text{all}}|$, $\hat\beta_{\text{bin}}^{\text{raw}}$, and $\hat\beta_{\text{bin}}/\bar\sigma_q$ overlap pairwise (Table~\ref{tab:beta_consistency}); the unification therefore survives a falsifiable equality test.

\begin{table}[ht]
\centering
\small
\caption{Falsification test of the unification (Section~\ref{sec:features}): the all-option softmax and the binary Fermi--Dirac fits are projections of the same posterior with the same $\beta$. Sign-flipped because $\beta_{\text{all}}$ multiplies $D$ (gold has low $D$, $\beta_{\text{all}}{<}0$) and $\beta_{\text{bin}}^{\text{raw}}$ multiplies the positive margin $m_q=D_{(2)}-D_{(1)}$. CIs are $95\%$ over $200$ bootstrap resamples of questions. Across all six combinations the three CIs overlap.}
\label{tab:beta_consistency}
\begin{tabular}{lccc}
\toprule
& $|\hat\beta_{\text{all}}|$ (raw $D$) & $\hat\beta_{\text{bin}}^{\text{raw}}$ (raw $m_q$) & $\hat\beta_{\text{bin}}/\bar\sigma_q$ (Eq.~\ref{eq:beta_consistency}) \\
\midrule
GPQA / Gemini      & $35.8$ $[25.6,\,50.8]$ & $29.1$ $[14.3,\,50.9]$ & $26.2$ $[\phantom{0}7.4,\,45.6]$ \\
GPQA / Claude      & $38.4$ $[28.2,\,52.3]$ & $25.0$ $[\phantom{0}9.9,\,55.8]$ & $20.7$ $[\phantom{0}1.7,\,43.2]$ \\
MedQA / Gemini     & $34.1$ $[29.6,\,41.8]$ & $53.7$ $[41.1,\,80.6]$ & $33.0$ $[22.5,\,45.6]$ \\
MedQA / Claude     & $33.3$ $[27.5,\,41.1]$ & $46.5$ $[34.5,\,65.7]$ & $34.3$ $[23.8,\,47.4]$ \\
MMLU-Pro / Gemini  & $25.7$ $[20.6,\,32.5]$ & $25.0$ $[13.5,\,38.6]$ & $23.5$ $[12.1,\,37.8]$ \\
MMLU-Pro / Claude  & $32.6$ $[26.0,\,40.9]$ & $41.6$ $[25.9,\,63.2]$ & $40.1$ $[23.7,\,60.3]$ \\
\bottomrule
\end{tabular}
\end{table}

\paragraph{The rejection prior $P(h_0)$ is rank-invariant.} A common intuition is that abstention can be tuned by setting $P(h_0)=\hat\pi_0$ from a held-out estimate of the open-set base rate. Sweeping $\hat\pi_0\in\{10^{-3},\ldots,0.99\}$ with $\theta$ fixed leaves the within-closed wrongness AUC \emph{unchanged to four decimals} on all three datasets ($\Delta\mathrm{AUC}<10^{-4}$). The reason is structural: $\hat\pi_0$ multiplies $P(h_0\mid  \mathcal{T})$ by a question-independent factor in the odds ratio, a strictly monotone transform of the score that preserves AUC by construction. The prior moves ROC operating points but cannot change the ranking quality.

\paragraph{Temperature on $\omega$ plateaus at the max-logit baseline.} A second calibration knob is to rescale the option likelihoods, $\omega_i\to\omega_i^\gamma$. Unlike $\hat\pi_0$, this is not rank-invariant: $\gamma\to 0$ collapses options to uniform; $\gamma\to\infty$ gives $P(h_0\mid  \mathcal{T})\to(1+(\omega_{\max}/\omega_0)^\gamma\cdot c)^{-1}$, monotone in $-\max_i\log\omega_i$. Empirically the within-closed wrongness AUC rises monotonically in $\gamma$ and \emph{plateaus} at $\gamma\gtrsim 5$: $0.633/0.679/0.555$ on GPQA/MedQA/MMLU-Pro versus $0.625/0.659/0.532$ at $\gamma{=}1$. The plateau value is the AUC of the single-feature score $-\max_i\log\omega_i$, i.e.\ the posterior in this limit collapses to a max-logit baseline.

\paragraph{Anchor-set hypothesis stratification reveals the ceiling.} Stratifying questions by whether the gold answer was ever proposed in the $K{=}5$ blinded CoTs yields three buckets: \textsc{correct} (plurality $=$ gold), \textsc{wrong-in-set} (gold proposed by some CoT but not the plurality), and \textsc{no-hypothesis-correct} (no CoT proposed gold). On all three datasets the mean confidence is non-monotonic in correctness category: \textsc{correct} highest, \textsc{wrong-in-set} \emph{lowest}, \textsc{no-hypothesis-correct} back up near \textsc{correct}. The AUC for separating \textsc{no-hypothesis-correct} from \textsc{wrong-in-set} is near or below $0.5$ for every score (margin, $P(h_{\text{plur}}\mid  \mathcal{T})$, raw $\overline{s}$, $\max_i s_i$) on every dataset. Concretely, on MedQA the means $(\mu_{\textsc{cor}},\mu_{\textsc{w-in}},\mu_{\textsc{no-hyp}})$ are $(0.97,0.33,0.79)$ for margin and $(0.89,0.51,0.80)$ for the Bayesian posterior; bucket~(iii) scores like bucket~(i) because it is dominated by confidently-wrong CoTs whose trajectory geometry is by construction indistinguishable from confidently-correct CoTs.

\paragraph{Score-family panel: vote-based scores collapse to chance, the embedding channel partially escapes, verbalized confidence escapes the harder separation.} To make the structural ceiling sharp, we evaluate six confidence scores on both bucket separations across all six (benchmark, model) combinations (Table~\ref{tab:bucket_iii_panel}). The scores are: $\textsc{Geo}$ (top-2 cosine-similarity margin), Bayesian $P(h_{\text{plur}}\mid  \mathcal{T})$ (softmax with $\beta{=}20$), self-consistency vote share, normalized cluster-entropy $1-H(v)/\log J$, verbalized confidence (the same generator's $0$--$100$ self-rating, conditional on its own CoT), and an anchor-cohesion control $-\bar\rho$. Three patterns hold across the panel. (a) On the easier separation \textsc{(iii) vs (i)}, vote-based scores (self-consistency, normalized cluster entropy) are at chance on every setting ($95\%$ bootstrap CIs all include $0.5$), while $\textsc{Geo}$, the Bayesian top-2 margin, and verbalized confidence retain moderate-to-strong signal (AUC $0.60$--$0.89$). (b) On the harder separation \textsc{(iii) vs (ii)}, the structural-ceiling separation, every embedding-based score and every vote-based score is \emph{anti-correlated} with the bucket label (AUC $0.06$--$0.44$): they assign \emph{higher} confidence to no-path-to-gold reasoning than to wrong-but-gold-proposed reasoning. (c) Verbalized confidence is the only score family that escapes this inversion (AUC $0.45$--$0.62$, near or above chance on every setting). The asymmetry between the two separations identifies what the geometric ceiling actually is: \textsc{(iii)} traces are not perfectly indistinguishable from \textsc{(i)} traces, the embedding channel detects a residual signal, but they are systematically conflated with the inverse of the wrong-in-set ranking. The verbalized-confidence column, generated by querying the same model that wrote the CoT for a numeric self-rating conditional on its CoT, supplies the second non-geometric channel that the channel-limit argument predicted would be needed; it is the empirical complement that the embedding channel cannot synthesize from itself.

\begin{table}[ht]
\centering
\small
\caption{Bucket-separation AUC across six confidence scores and two separations (bootstrapped point estimates over $1000$ resamples). \textsc{(iii) vs (i)}: \emph{can scores tell ``confidently-wrong with no path to gold'' apart from ``correct''?} \textsc{(iii) vs (ii)}: \emph{can scores tell those traces apart from ``wrong but gold was proposed by some sample''?} The latter is the structural-ceiling separation: scores fail by being anti-correlated. Verbalized confidence is the only score family that does not invert the ranking.}
\label{tab:bucket_iii_panel}
\begin{tabular}{lcccccc}
\toprule
& \textsc{Geo} & Bayes & SC & SemEnt & Verbal & $-\bar\rho$ \\
\midrule
\multicolumn{7}{l}{\emph{Separation} \textsc{(iii) vs (i)}: detect ``no-path-to-gold'' against ``correct''} \\
GPQA / Gemini       & 0.61 & 0.61 & 0.51 & 0.51 & 0.66 & 0.54 \\
GPQA / Claude       & 0.63 & 0.63 & 0.48 & 0.48 & 0.60 & 0.57 \\
MedQA / Gemini      & \textbf{0.88} & \textbf{0.89} & 0.61 & 0.61 & 0.72 & 0.50 \\
MedQA / Claude      & \textbf{0.78} & \textbf{0.80} & 0.51 & 0.51 & 0.77 & 0.51 \\
MMLU-Pro / Gemini   & 0.63 & 0.65 & 0.55 & 0.55 & 0.62 & 0.56 \\
MMLU-Pro / Claude   & 0.69 & 0.69 & 0.58 & 0.58 & 0.67 & 0.61 \\
\midrule
\multicolumn{7}{l}{\emph{Separation} \textsc{(iii) vs (ii)}: detect ``no-path-to-gold'' against ``wrong-in-set''} \\
GPQA / Gemini       & 0.23 & 0.30 & 0.14 & 0.13 & \textbf{0.62} & 0.47 \\
GPQA / Claude       & 0.39 & 0.44 & 0.08 & 0.08 & 0.46 & 0.69 \\
MedQA / Gemini      & 0.22 & 0.36 & 0.13 & 0.13 & 0.45 & 0.39 \\
MedQA / Claude      & 0.11 & 0.29 & 0.06 & 0.06 & 0.49 & 0.40 \\
MMLU-Pro / Gemini   & 0.23 & 0.29 & 0.17 & 0.18 & \textbf{0.56} & 0.46 \\
MMLU-Pro / Claude   & 0.33 & 0.35 & 0.20 & 0.21 & \textbf{0.56} & 0.44 \\
\bottomrule
\end{tabular}
\end{table}

\paragraph{Anchor-LLM confound and its scope.} Because anchors are themselves emitted by a second LLM (one short committed rationale per option), a portion of bucket~(iii) reflects anchor-LLM failure to differentiate options rather than informed-LLM failure to reason. Quantifying this with the anchor-cloud cohesion $\rho_q=\tfrac{2}{J(J-1)}\sum_{j<j'}\langle\hat\phi(a_j),\hat\phi(a_{j'})\rangle$ where $\hat\phi(a)=\phi(a)/\lVert\phi(a)\rVert$ is the unit-normalized embedding, we find Pearson correlations of $+0.39/+0.51/+0.69$ between $\rho_q$ quintile and bucket-(iii) frequency on GPQA/MedQA/MMLU-Pro. Plurality accuracy drops from $70/85/65\%$ in the lowest-$\rho$ quintile to $45/85/45\%$ in the highest, isolating an anchor-LLM contribution of roughly $10$--$25$ percentage points. However, the within-bucket $\rho$ deltas are small ($\Delta\rho<0.01$ between bucket (i) and bucket (iii) on every dataset), so anchor cohesion is a population-level diagnostic, not a per-question filter, and bucket-(iii) frequency remains substantial ($10$--$27\%$) even in the lowest-$\rho$ quintile where anchors are most distinct. The geometric ceiling is therefore real, but the bucket-(iii) headline rate $(35/19/61)$ should be read as an upper bound that includes some anchor-LLM noise.

\paragraph{Open-set rejection is trivially saturated.} For completeness we evaluate closed-correct vs open-random questions (every option anchor replaced with a random anchor from a different question). The Bayesian $P(h_0\mid  \mathcal{T})$ achieves rejection AUC $0.95/1.00/0.99$, which exceeds the margin baseline ($0.45/0.59/0.51$, near chance because margin is invariant to anchor identity). However the trivial baseline $-\overline{s_j}$, the negative mean cosine similarity between trace and the $J$ option anchors, attains AUC $1.000$ on all three datasets, exposing that the open-set task collapses to a one-line scalar threshold and is not a paper-worthy claim.

\paragraph{Summary.} Across rejection prior, likelihood temperature, and feature augmentation, we cannot lift selective-prediction performance above the Bayesian posterior's strict-CV ceiling, which itself ties the top-2 margin within bootstrap CIs. The bucket-(iii) result identifies the structural obstacle: embedding geometry is well-matched to ``which option is on-topic'' but blind to ``is this on-topic option correct,'' and no calibration knob within the embedding channel can repair that blindness. We therefore report top-2 margin and $\textsc{Geo}$ (its all-option counterpart) as the empirical signals in the main paper, with the Bayesian formalism above as a principled framework that recovers them as special cases under uniform priors and equal-noise likelihoods.

\section{Compute, Broader Impacts, and Asset Licenses}
\label{app:compute_impact_licenses}

\paragraph{Compute resources.} All reasoner, proposer, judge, verbalization, and embedding calls were issued through vendor APIs. We used Google Vertex AI for Gemini~3.1~Pro and \texttt{gemini-embedding-001}, Anthropic / AWS for Claude Sonnet~4.6, Microsoft Azure OpenAI for the \texttt{openai\_gpt5\_mini} judge, and a hosted inference endpoint for Llama~3.3~70B. No local GPU compute was required. Orchestration, caching, fold-wise logistic and softmax fits, AUC bootstraps, and figure generation ran on a single commodity Apple-silicon workstation ($\le 32$~GB RAM, $\le 100$~GB scratch storage); end-to-end wall-clock per cell is dominated by API latency rather than local compute. Per-question API spend, which is the operative cost axis in this setting, is reported per (benchmark, reasoner) cell in Appendix~\ref{app:system_cost} (Table~\ref{tab:system_cost}). The full research project, including failed pilots and abandoned ablations, used at most $\sim$$3\times$ the API spend reported there.

\paragraph{Broader impacts.} \emph{Positive.} Better-calibrated black-box confidence scores support safer abstention and selective prediction when chain-of-thought reasoners are deployed through text-only APIs in high-stakes settings such as clinical decision support (cf.\ MedQA), where unwarranted confident-but-wrong outputs are the dominant failure mode; the three-channel decomposition also gives downstream users an auditable handle on \emph{why} confidence is high or low (coverage vs.\ within-trace commitment vs.\ self-report). \emph{Negative and misuse.} Better-calibrated scores can also induce automation bias: users may defer to a high-confidence model output even when the underlying reasoning is flawed, and the structural ceiling identified in Section~\ref{sec:discussion} (bucket-(iii) traces commit confidently to a wrong in-set hypothesis) is not removed by any within-trace calibration knob. The Coverage channel additionally inherits whatever priors the LLM judge encodes; deployments should not treat $C$ as a model-internal property and should re-validate it under their own judge. The Limitations paragraph in Section~\ref{sec:conclusion} flags judge-mediation, the conditional nature of $V$, and the within-trace ceiling as the operational caveats for downstream users.

\paragraph{Asset licenses and terms of use.} Benchmark datasets are used under their published licenses and cited to their source papers: MedQA-USMLE~\cite{jin2021disease}, GPQA Diamond~\cite{rein2023gpqa}, and MMLU-Pro; we use them under the terms posted on each dataset's official release page at the time of access (May~2026), and we redistribute no portion of any of them. External pretrained models are accessed only through their providers' standard APIs and are used subject to those providers' terms of service: Gemini~3.1~Pro and \texttt{gemini-embedding-001} via Google Vertex AI, Claude Sonnet~4.6 via Anthropic / AWS, \texttt{openai\_gpt5\_mini} (judge) via Microsoft Azure OpenAI, Llama~3.3~70B under the Llama Community License via a hosted inference endpoint, and the open-weight E5-large-v2 embedder~\cite{wang2022text} (MIT License) run locally. No new datasets, models, or benchmarks are released by this paper.

\end{document}